\documentclass{article}

\usepackage[margin=1in]{geometry}
\usepackage{graphicx}
\usepackage{siunitx}
\usepackage{subcaption}
\usepackage{caption}
\usepackage{longtable}
\usepackage{lineno,hyperref}
\usepackage{enumitem}
\usepackage{authblk}
\usepackage{pdflscape}

\title{Perceiving University Students' Opinions from Google App Reviews}
\date{2013-09-01}
\author[1]{Sakshi Ranjan \thanks{Corresponding Author \\ Accepted at Concurrency and Computation Practice and Experience }} 
\author[2]{Subhankar Mishra}

\affil[1]{Department of Computer Science, Utkal University, Bhubaneswar-751004, India. sakshi.ranjan07@gmail.com}
\affil[2]{School of Computer Sciences, National Institute of Science Education Research, Bhubaneswar-752050, India. \\ Homi Bhabha National Institute, Anushaktinagar, Mumbai - 400094, India. smishra@niser.ac.in}

\begin{document}
  \maketitle

\begin{abstract}
Google app market captures the school of thought of users from every corner of the globe via ratings and text reviews, in a multi-linguistic arena. The critique’s viewpoint regarding an app is proportional to their satisfaction level. The potential information from the reviews can’t be extracted manually, due to its exponential growth. So, Sentiment analysis, by machine learning and deep learning algorithms employing NLP, explicitly uncovers and interprets the emotions. This study performs the sentiment classification of the app reviews and identifies the university students’ behavior towards the app market via exploratory analysis. We applied machine learning algorithms using the TP, TF and, TF-IDF text representation scheme and evaluated its performance on Bagging, an ensemble learning method. We used word embedding, GloVe, on the deep learning paradigms. Our model was trained on Google app reviews and tested on Students’ App Reviews(SAR). The various combinations of these algorithms were compared amongst each other using F-score and accuracy and inferences were highlighted graphically. SVM, amongst other classifiers, gave fruitful accuracy(93.41\%), F-score(0.89) on bi-gram+TF-IDF scheme. Bagging enhanced the performance of LR and NB with accuracy 87.88\% and 86.69\% and F-score 0.86 and 0.78 respectively. Overall, LSTM on Glove embedding recorded the highest accuracy(95.2\%) and F-score(0.88).

\end{abstract}
  
  \section{Introduction}
We are living in an era where technology and the Internet have redefined social norms. There is no denying that mobile apps have changed every aspect of our lives completely\cite{M.Harman2012}. Irrespective of what we want or need to do; everything is at our fingertips, just by discovering the relevant apps and reading the reviews and ratings posted by others. This helps in generating profit for the developers, giving bug reports, fierce competition amongst apps, requests for new features, documentation of experience to analysts\cite{L.V.Galvis2013} and designers\cite{W.Maleej2011}. It gives information related to products, services, organizations, individual issues, events, satisfaction or dissatisfaction with new features, or business relevant information to software developers. 
Whether we are traveling\cite{Y.Blanco2010}, communicating\cite{P.Adinolfi2016}, watching movies\cite{T.T.Thet2010}, ordering products\cite{H.Cui2006}, performing bank transactions, there is an app for everything, and so is the review. The proliferation of Google apps has helped us realize the rich interplay concerning the users-, trading-, and technologically concentrated traits\cite{Li}.

The motivation for studying the Google app reviews and conducting this study is as follows.
In the past decades, smartphones were uncommon so there were fewer interactions worldwide. Mainly the source of information was news, the Internet, and other sources of media. 
Internet, mobile technology, and networking infrastructures have brought the inception and explosive growth of Google Play store and Apple store apps into being (like Facebook, Twitter, Instagram, Kindle, Amazon, Google pay, etc.) \cite{A.Go, ss}. People write and publicize their reviews and ratings from across the globe, based on the apps on their devices, satisfaction rate, and likings\cite{Kohli}. These have coincided with social media on the Web (reviews, discussions forum, blogs, micro-blogs, Twitter) and provided us with rich sources of data for researching\cite{Chen}.
Google Play store app market captures countless responses per month. 
Using sentiment analysis and mechanizing this process\cite{Chua}, we can benchmark how users feel about apps without having to read thousands of user comments at once\cite{F.Tian}.
Particularly, just by studying the ratings for a given app, the criteria of understanding the mindset of a person, cannot be fulfilled. This is because ratings do not provide tangible statistics. So studying and analyzing the real-time reviews is also a necessity. 

New apps are rolling out every day with technical and multifaceted information available in the description; and ordered in terms of the latest reviews, ratings, download strategy\cite{N.Seyff2010}. This helps in the qualitative and quantitative analysis of users' viewpoints for sizing and pricing strategy, technical claims, and features of apps. 
Natural Language Processing(NLP), a buzzword in recent research, mines the technical information from reviews.
It is one of the trending applications of Artificial Intelligence(AI) and interestingly comprehends the features. Many pioneer researchers are exploiting algorithmic approaches to understand the relationship between the claimed features\cite{Malik}.

However, the problem with the app market is its abundance of reviews that take extra effort and longer time in manual computations. One of the bottlenecks is the information overload problem and its noisy nature\cite{Liu}. Secondly, the quality of reviews varies tremendously from essential and innovative advice to offensive comments. Thirdly, filtering the negative and positive comments in the reviews and extracting feedback from them is sometimes tricky. Also, the unstructured nature of reviews is troublesome to parse and analyze. This study only focuses on English reviews given by users and not on multilingual sentiment analysis or resource-poor languages.


Sentiment analysis helps to mine the people’s opinions, sentiments, behaviors, emotions, appraisals, and attitudes towards products or services, issues or events, topics\cite{Liu2015}. There are three types of people's opinions namely, positive, negative, and neutral which identify the entire knowledge of the domain. 
It is an integral part of the NLP and enables text mining and information retrieval\cite{E.Fersini}. In the field of education, sentiment analysis refines the international education institutions by e-learning techniques\cite{C.L.Santos} and perceptions\cite{W.Y.Hwang}.
In recent years, it has extended to fields like marketing, finance, political science, communications, health science, education using a coherent framework\cite{P.Pang}. 
We can extract opinions using sentiment analysis tools, process the results and, come to valuable conclusions\cite{L.V.Galvis2013}. The resulting model from this study sets a new state-of-the-art to focus only on a bunch of university students and crawl their reviews regarding the play store apps they use and using NLP to introspect the sentiment associated with it\cite{Mudambi}.

Machine Learning-based techniques\cite{A.B}, as well as lexicon-based methods, are used in sentiment analysis\cite{W.Medhat2014}. Lexicon-based approach is an approach that considers the semantic order of the words and doesn't include labeled data. Dictionary is created manually and includes words and phrases in a document\cite{Godbole}. 
Sentiment analysis through a machine learning approach deals with labeled data and helps to create models using supervised learning algorithms namely, Naïve Bayes(NB), Support Vector Machine(SVM), and K‐nearest neighbor(KNN)\cite{A.Onan2016}. 
The Deep learning paradigm\cite{R.Prabowo}, an interdisciplinary of machine learning algorithms, based on fine-tuned layers, has outperformed major classification algorithms\cite{L.Deng}. It has yielded fruitful results in speech recognition, computer vision, and sentiment analysis\cite{a}. 
When used with word embedding, it scales well with fine-grained opinions and tunes itself with the hyper parameters\cite{X.Glorot}.

In our study, we had collected 10,841 Google app reviews with 13 fields to train our model\cite{data}. While for the sake of testing our model, we collected 400 reviews with 6 fields from amongst the Utkal university students via local survey, department-wise. 
Specifically, this paper presents the correlation between Students App Reviews(SAR) and the Google app reviews via an exploratory analysis and visualization of sentiment polarity, subjectivity versus other features like price, installs, type, size, category, ratings.
Towards this end, we initiate a methodical approach to mine opinions from Google app reviews and hence the contribution of our paper includes:
\begin{itemize}
    \item Several Research Questions(RQ) were designed and evaluated on the corpus through visualization using charts and making intuitive judgments.
    \item The text representation scheme namely, TP(Term Presence), TF(Term Frequency) and, TF-IDF(Term Frequency-Inverse Document Frequency) were implemented on uni-gram, bi-gram and, tri-gram strategies.
    \item The supervised machine learning methods(such as NB, SVM, logistic regression(LR), KNN, and Random Forest(RF)) were implemented on the text representation scheme and compared amongst each other concerning for its performance metrics.
    \item The ensemble learning method(namely, bagging) was used with the classification algorithm namely, LR and, NB and its performance was evaluated on the text representation scheme. 
    \item Fine-tuned Deep learning models like Long Short Term Memory(LSTM), Convolution Neural Network(CNN), Recurrent Neural Network(RNN) were implemented layer by layer on word embedding(GloVe) and its performance was noted and compared graphically.
\end{itemize}

The organization of this paper comprises five sections. Section 2 throws light on the related works in sentiment analysis. Section 3 describes the methods utilized in the paper. Section 4 highlights the empirical analysis, results. Finally, Section 5 wraps up with conclusions of the study and future scope in this context.

\section{Background}


\subsection{Related Works in Sentiment Analysis using Machine Learning approaches}

\begin{enumerate}
    
\item Lima et al.\cite{Lima2015} have used a majority voting scheme on the Twitter dataset. They have combined machine learning-based paradigms and lexicon-based methods. In their work, the tweets are a part of the labeled training data only when it consists of 5\% of words or emoticons, otherwise, it is considered a part of test data. Novak et al.\cite{Novak2015} have explained about emoji-based sentiment analysis and the 750 frequently used emojis were also analyzed in the Twitter dataset. Lately, Onan et al.\cite{Onan2020} have collected instructor reviews from students for opinion mining using machine learning and the deep learning paradigm. A comparison between different machine learning and deep learning algorithms was made and the inference was that GloVe with Recurrent Neural Network - Attention Mechanism(RNN-AM) algorithm has outperformed others.

\item While Adekitan and Noma‐Osaghae\cite{Adekitan2019} have predicted the performance of university students using machine learning algorithms in their work. Linear and quadratic regression models were used for validation. Almasri et al.\cite{Almasri2019} have predicted the performance of students using ensemble tree‐based models. While Adinolfi et al.\cite{P.Adinolfi2016} evaluated student satisfaction on different learning e-platforms of online courses using sentiment analysis.


\item Farhan et al.\cite{Farhan} proposed a research paper to mine opinions from Twitter data. The performed pre-processing of reviews for sentiment analysis. These include slang and abbreviation identification, correcting spellings, removing stop words, tokenization, stemming and, lemmatization. Emoticon identification was also done. They used lexicon-based approaches. SentiWordNet was used. Misclassifications of tweets were also handled efficiently 

\item Harman et al\cite{Harman} proposed that App Store Analysis can be used to understand the relation between the user, technical, market, and social aspects of app stores. They extended their study to the non-free app in the Blackberry app market. Also find the correlation between the claimed features, ratings, price, size, downloads. Feature extraction was done from the app descriptions.

\item Mcllroy et al.\cite{Mcllroy} have studied the updates strategy of mobile apps in Google play store apps. They inferred that 1\% of apps are updated weekly while  14\% of apps are updated very often. Ranking of frequently updated apps is done based on the frequency. New updates are not highlighted in 45\% of the frequently updated apps. 

\item Onan et. al. \cite{a, ss} used ensemble learning techniques for sentiment analysis involving feature engineering. They incorporated Bayesian logistic regression, naïve Bayes, linear discriminant analysis, logistic regression, and support vector machines as base learners and concluded that the laptop dataset showed the best accuracy of 98.86\%. Similarly, Turkish sentiment analysis was done using nine supervised and unsupervised term weighting schemes by Onan\cite{ b}. He inferred  that supervised term weighting methods gave fruitful results over unsupervised term weighting methods. Onan et.al.\cite{c} presented a work where an ensemble approach for feature selection was aggregated with the different individual feature lists obtained by several feature selection methods.

\item Onan et. al.\cite{j} examined the performance of five statistical keyword extraction methods using classification algorithms and ensemble methods for text classification. Bagging with Random forest turned out best for their study(93.8\%) and could have practical applications too. Onan et.al.\cite{m} presented ensemble methods for satirical news identification in Turkish news. Supervised learning methods along with ensemble methods were used. They inferred that random forest(96.92\%) and recurrent neural network with attention mechanism( 97.72\%) were best for their study.

\item Onan\cite{l}  presented a comparative analysis of several feature engineering schemes and classification algorithms aggregated with ensemble methods. Random subspace is used with random forest using four features. An accuracy of 94.43\% was obtained for the corpus.
An ensemble scheme using hybrid supervised clustering  for text classification is shown in the work of Onan\cite{n}.  Supervised hybrid clustering method based on cuckoo search and the k-means algorithm was used on the corpus for clustering and the results were compared with conventional classification algorithms. In another instance Onan et. al.\cite{o} have shown that  hybrid ensemble pruning schemes with clustering and randomized searches can yield fruitful results in the field of text classification. They presented a consensus clustering scheme also. They inferred that the consensus clustering and the elitist pareto-based multi-objective evolutionary algorithm can be effectively employed in ensemble pruning.




\end{enumerate}

\subsection{Related Works in Sentiment Analysis using Deep Learning approaches}

\begin{enumerate}
    \item Deep learning techniques have been used for opinion mining and emotion recognition and used for educational tasks employing data mining. In a study, Bustillos et al\cite{Oramas} examined supervised algorithms and LSTM and CNN algorithms showing an accuracy of 88.26\%. Similarly, Cabada et al\cite{Cabada} showed deep learning architectures for sentiment analysis based on the educational system. The  emphasis was on CNN architecture and LSTM attained an accuracy of 84.32\%.

\item A comparative survey between machine learning and deep learning algorithms was presented by Sultana et al\cite{Sultana} on educational data. In their study, they showed the highest predictive performance was claimed by SVMs and multi-layer perceptrons with an accuracy of 78.75\% and 78.33\%, respectively.

\item In another instance, Nguyen et al\cite{Nyugen} emphasized Vietnamese students' reviews using machine learning and deep learning techniques. NB was used; LSTM and bidirectional LSTM  were included in empirical analysis. Unigram and bigram features were calculated for the corpus while the word2vec word embedding scheme was calculated on deep learning algorithms. They got a crystal clear inference that deep learning‐based architectures yield higher predictive performance in comparison with conventional
machine learning classifiers. Bidirectional LSTM showed an accuracy of 89.3\%

\item Zhou et al\cite{Zhou} assessed the sentiment analysis of movie reviews using the Stanford Sentiment Treebank(SST) corpus by employing deep learning techniques. They inferred that CNN and LSTM outperformed CNN and RNN models. Positive negative(2-class) reviews achieved 87.7\% accuracy while the 5-class reviews(very positive, positive, neutral, negative, very negative) attained 49.2\% accuracy.
When GloVe was used on the corpus, accuracy had risen to 88\%.

\item Zhang et al.\cite{Zhang} proposed a sentence-level neural model approach to overcome the weakness of pooling functions that don't uncover tweet-level semantics. They used two gated neural networks, namely a bi-directional gated neural network and a three-way gated neural network to model the interaction between target text and surrounding contexts. The bias of RNN is also reduced. Moreover, words were connected in the tweets to apply pooling functions over the hidden layers of texts.      


\item Kandhro et al.\cite{Kand} used LSTM for analysis of the sentiments expressed by students through reviews for their teachers. The corpus used for this study was built through student feedback and then divided into 70\% and 30\% for training and testing purposes. They inferred that the model's accuracy was 99\% and 90\% for training and the loss was 0.2 and 0.5 respectively during validation. Their model overcomes the issues of Bag of words, SVM and, Naive Bayes.

\item Onan et al.\cite{e} have presented a sarcasm identification framework using social media texts in which sarcastic text documents were modeled using 3-layered stacked bidirectional LSTM architecture and got an accuracy of 95\%. They also evaluated
three neural language models, two unsupervised term weighting functions. In another instance, Onan\cite{f} has presented an efficient sentiment classification scheme for Massive open online courses(MOOC) reviews, using the ensemble learning and deep learning methods. He inferred long short‐term memory networks in conjunction with GloVe word‐embedding scheme‐based representation, and got an accuracy of 95.80

\item Onan\cite{d} used the product reviews obtained from Twitter for sentiment analysis. CNN-LSTM approach was used with TF-IDF and GloVe schemes. Although the conventional deep neural networks were used to assess the performance, deep learning approaches outperformed others. Similarly, Onan\cite{i} has presented a deep learning approach for sarcasm analysis. Six subsets of twitter dataset have been considered ranging from 5000 to 30000. The topic-rich word embeddings gave fruitful results.

\end{enumerate}

\subsection{Observations from Literature Survey}
The use of conventional text representation schemes with machine learning algorithms and also deep learning algorithms has drawn research attention, lately\cite{c}\cite{g}\cite{h}.
However, according to our study, there are very limited works based on predictive performances of algorithms using Google app reviews in conjunction with university students' reviews related to Google apps. There are no past reports on NLP in sentiment analysis of user reviews regarding the Google apps in conjunction with SAR.
Precisely, the state-of-the-art methods do not capture any of the comparisons between the machine learning algorithms and deep learning algorithms. 
And to the best of our knowledge, the combination of techniques used in our study is a bit unique.
Table- \ref{t1} captures a comparison of existing literature in context with Sentiment analysis based on different reviews crawled in different languages and domains.

For bridging this gap, our literature survey was inspired by instructors' review paper approach\cite{Onan2020} that threw light on multiple combinations of machine and deep learning algorithms. The latest trend observed from our work is that it does not emphasize the count vectorizer method of splitting the data set rather aggregates a new university data set. We incorporated data analysis along with modeling. Moreover, basic research questions, in context with the domain of Google app reviews and SAR were answered via charts.

\begin{table}[ht]
\centering
\caption{Comparison between Existing Literature of NLP for Instructor Reviews \cite{Onan2020}}
\begin{tabular}[t]{llr}
\hline

Reference & Methods & Accuracy\\
\hline
Sultana et al\cite{Sultana}    &  Multilayer perceptron  &             78.33  \\            
Sultana et al\cite{Sultana}  &Support vector machines   &         78.75\\
Nguyen et al\cite{Nyugen}   &    Unigram features + Naive Bayes  &   85.30\\
Nguyen et al\cite{Nyugen}     &    Bigram features + Naive Bayes &   87.50\\
Nguyen et al\cite{Nyugen}     &   word2vec + LSTM              &     87.60\\
Nguyen et al\cite{Nyugen}      & word2vec + Bi‐LSTM           &     92.00\\
Kandhro et al\cite{Kand}  &word2vec + LSTM        &           89.00\\
Bustillos et al\cite{Oramas}   & Bernoulli Naive Bayes &            76.77\\
Bustillos et al\cite{Oramas}   &CNN + LSTM  &                     88.26\\
Cabada et al\cite{Cabada}   &     Multilayer perceptron &            90.42\\
Cabada et al\cite{Cabada}    &    CNN              &                 92.46\\
Cabada et al\cite{Cabada}     &   LSTM            &                   90.92\\
Cabada et al\cite{Cabada}     &  CNN + LSTM     &                    92.15\\
Onan et al\cite{Onan2020}&            Glove + RNN‐AM  &                  98.29\\
\hline
\end{tabular}
\label{t1}
\end{table}%

\begin{table}[ht]
\centering
\caption{Sample students’ reviews and Sentiment Characteristics from Students’ dataset.}
\label{t2}
\begin{tabular}{p{0.45\textwidth}llrr}
\hline
Student's Reviews & Apps & Orientation & Polarity & Subjectivity  \\ \hline
It’s helpful to learn at home Highly recommendable. & Unacademy         & Positive & 0.04   & 0.135 \\ 
It's amazing and works well.                        & Phone pay         & Positive & 0.3    & 0.725 \\
Horrible. Keeps crashing my phone                   & Subway Surfers    & Negative & -0.104 & 0.43  \\
It's annoying due to adds.                          & JioSaavn          & Negative & -0.033 & 0.388 \\
Very well designed. Many apps present.              & WPS Office        & Positive & 1      & 0.75  
 \\ \hline
\end{tabular}
\end{table}

%

\section{Data Sources and Methodology}
\begin{figure}
    \centering
    \includegraphics[width=0.8\textwidth]{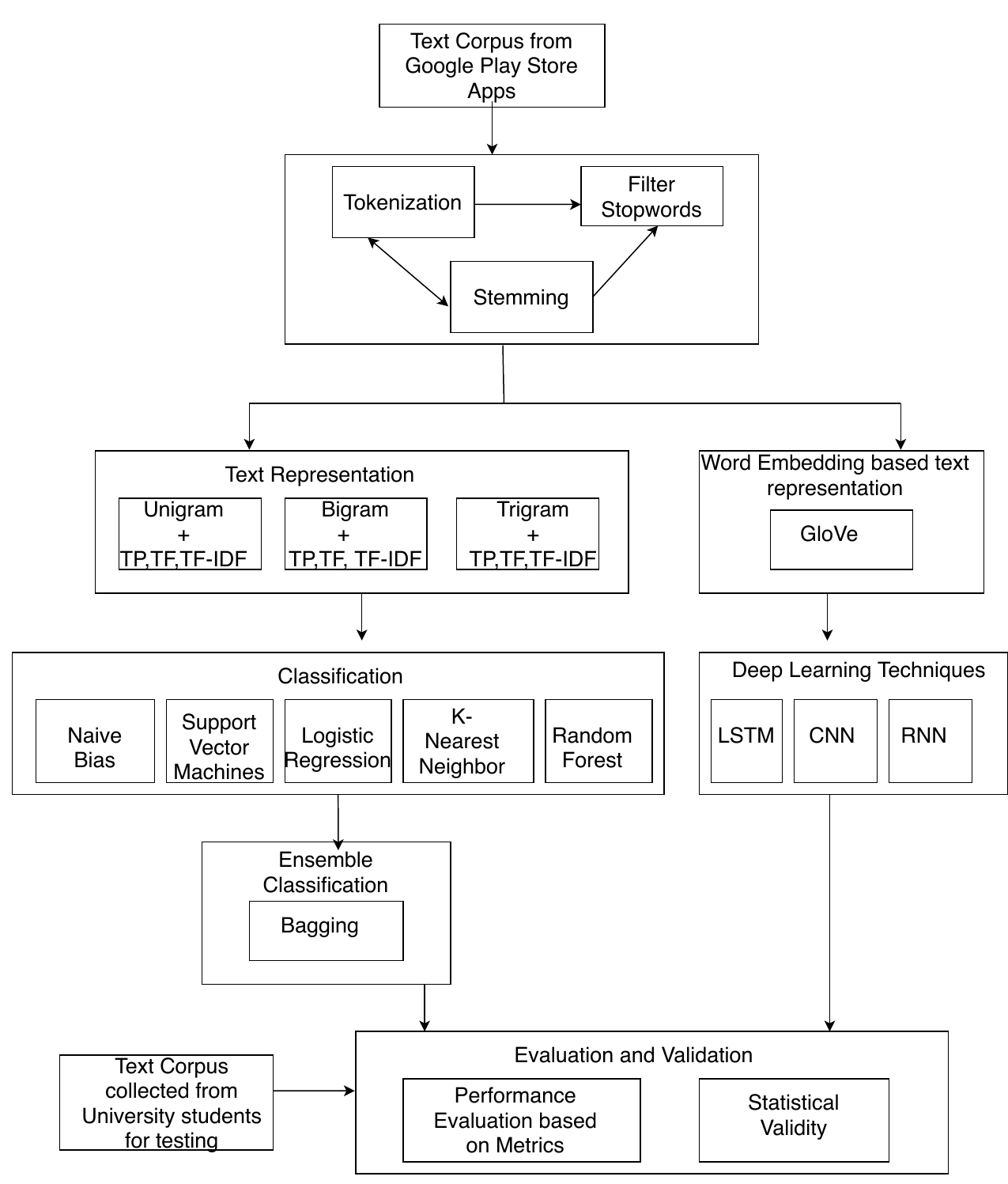}
    \caption{Architectural Framework for Sentiment Analysis.}
    \label{f1}
\end{figure}

Fig.\ref{f1} explains the proposed methodology of our study. 

\subsection{Data Sources}
This study had 2 data sets under consideration and are listed below:
\begin{itemize}
    \item Training data set, Google app reviews.
    \item Test data set, SAR.
\end{itemize}

\textbf{Google app reviews}- The corpus is openly available for research and it was collected in .csv format\cite{data}. There were 9659 apps, 33 categories, 115 genres in the dataset. The columns of the data set are as follows app(name), category(app), rating(app), reviews(user), size(app), installs(app), type(free/paid), price(app), content rating(everyone/ teenager/ adult), genres(detailed category), last updated(app), current version(app), android version(support).

\textbf{SAR} - In this study, the aim is to understand the trend of the Google app market and comparing it with test data i.e., analyzing the students' behavior towards the Google app market. 
So, we had collected the real-life data from the Utkal university students, department-wise. 
The survey was entirely voluntary and no incentives were offered to perform the survey. If university students did not wish to participate, then they were excluded from the survey.
The reviews regarding the frequently used apps were gathered via a survey for a month. The survey was made on an online platform via a Google form. One student from one department could at max list out seven frequently used apps on their device. They had given their reviews in English. There were 400 data collected from the survey with 6 fields including department (name), app (name), reviews (user), ratings(everyone/teenager/adult), type(free/paid), category. Students rated the apps on a 5-point scale where ratings below 3 were labeled as "negative" and that with 3 or greater than 3 is considered "positive". 
Furthermore, in the empirical analysis data cleaning, text pre-processing techniques were carried out on the dataset to build efficient learning models and enhance the overall performance. These include removing  missing data, dropping NA values, removing punctuation, tags, special characters URLs, emojis, digits, filtering stop words, tokenization, noise removal, spelling correction, stemming, and lemmatization\cite{GAMiller1995} 
After cleaning, there were 9,360 and 380 records in the training and test dataset respectively.
Table-2 presents some sampled test data reviews along with the sentiment characteristics. 
Orientation - determines positivity, negativity, or neutrality of sentence, Polarity - helps identify the sentiment orientation, Subjectivity - defines person' opinions, emotions, or judgment; ranging from 0.0 (objective) to 1.0 (subjective). 
A sentiment score determines how negative or positive the entire text analyzed is. 
For eg., the phrase “not a very great app” has a polarity of about -0.3, which implies it is slightly negative, and  subjectivity of about 0.6, implies it is fairly subjective.

\subsection{Text Representation Schemes}

One of the motives to study the text representation schemes(namely, Bag-of-words, TP, TF and, TF-IDF) is feature engineering, used especially for NLP applications. Features act as input parameters for the machine learning algorithms to generate some output and enabled the classifiers to model its performance. 

The Bag-of-words paradigm\cite{G.Hackeling2017} is a very commonly used technique to represent all the unique words occurring in the documents. The occurrences of the terms in a document are noted while the order and the sequence of words are not considered. 
The three weighted schemes frequently utilized are based on the bag-of-words model i.e., TP, TF and, TF-IDF. 
 TP maps the appearance of words in a document, by binary values 1 and 0, indicating their presence and absence respectively.
 TF counts the number of appearances of words in a document. Commonly used words have a higher count in context with rarely appeared words.
The TF-IDF scheme is an improvement over TP\cite{J.Brownlee} and uses a normalizing aspect for computations. Mathematically, TF-IDF is defined as: 
\begin{equation}
TF-IDF = TF(w,D)* log(C/df(w))
\end{equation}




N-gram model is a collection of words from a text document in which the words are contiguous and occur sequentially. They may be in the form of phrases or groups of words.
In this study, we performed an experiment on the Google apps corpus and SAR based on three N-gram model(unigram-it consist of one word, bigram-it consists of two words and trigram-indicates n is three) and TP, TF, TF-IDF, and obtained nine different configurations.

\subsection{Ensemble Learning Methods}

The base estimators are built on a given learning algorithm and their predictions can be combined to improve the robustness and performance over a single estimator\cite{A.Onan2017}. It includes averaging and boosting methods. 
\textbf{Bagging} or Bootstrapped Aggregation is used for predictive modeling (CART)\cite{L.Brieman2001}. Random subsets of data are drawn from the training dataset with replacement, and a final model is produced by averaging results from several models\cite{b}.
One popular way of building Bagging models is by combining several DecisionTrees with reduced bias that increases the model’s prediction than individual Decision Trees\cite{j}. Averaging ensembles with bagging techniques like RandomForestClassifier and ExtraTreesClassifier reduces the variance, avoids over-fitting, and increases the model's robustness concerning small changes in the data\cite{l} \cite{n} \cite{o}.

\subsection{Word Embedding}
One of the importance of word embedding is to convert the text into vector representation(numerical format) using some statistics as computers can't understand natural language directly. 
Word embeddings are dense vector representations with lower dimensionality and overcome word ambiguities\cite{S.M2019}.
It provides an improvement over the simplest bag-of-words model, widely used in NLP\cite{d} \cite{i}. 
From our study, we can infer word embeddings outperforms conventional text representation schemes in NLP. 



{\textbf{GloVe}}- One of the necessities to study GloVe is to focus on the distributional semantics(develop theories that quantify and categorize semantic similarities between linguistic items based on the distributional properties in large samples). GloVe overcomes the shortcomings of one hot vector encoding, FastText and word2vec\cite{Penni}. 
 For instance, in the research of Onan et al.\cite{Onan2020} using instructors' reviews from students for text analysis, GloVe was used with all deep learning paradigms(LSTM, RNN, CNN, RNN-AM, GRU). Upon comparison, GloVe with RNN - AM algorithm has outperformed others, this urged us to handpick Glove as one of the architectural components in our study.
 




\subsection{Algorithms}
\begin{figure}
     \centering
     \begin{subfigure}[b]{0.45\textwidth}
         \centering
         \includegraphics[width=\textwidth]{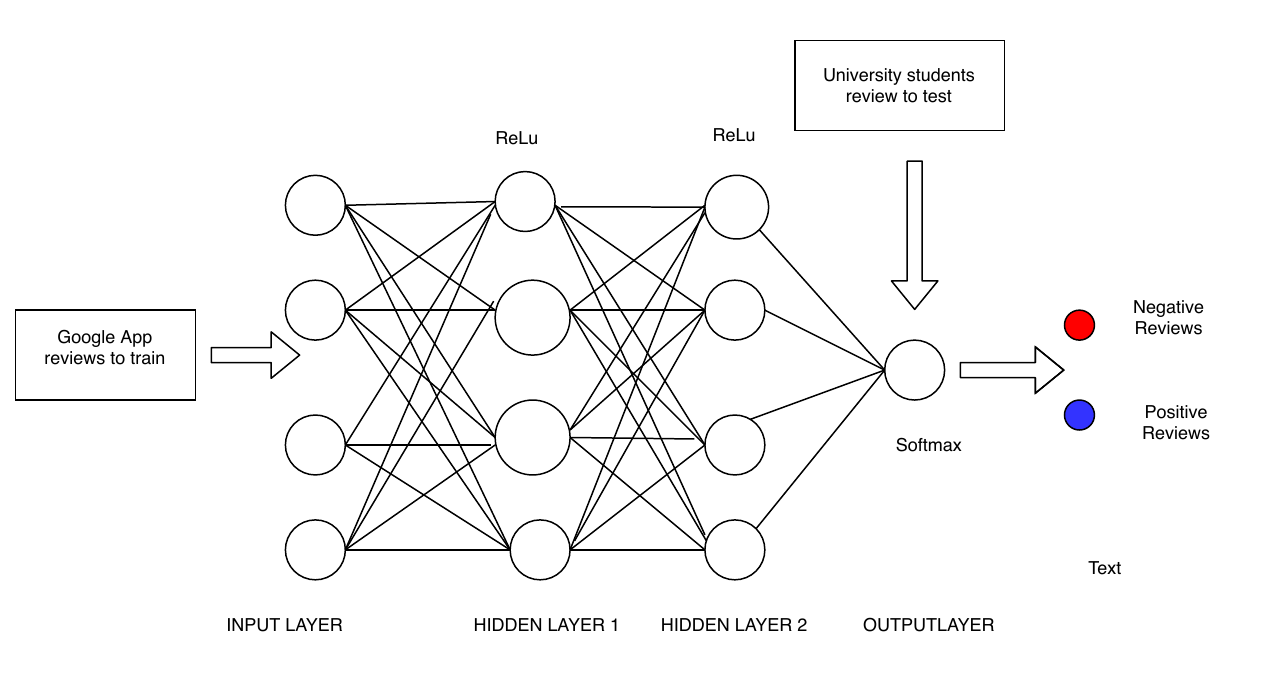}
         \caption{MLP Architecture.}
         \label{f2}
     \end{subfigure}
     \hfill
     \begin{subfigure}[b]{0.45\textwidth}
         \centering
         \includegraphics[width=\textwidth]{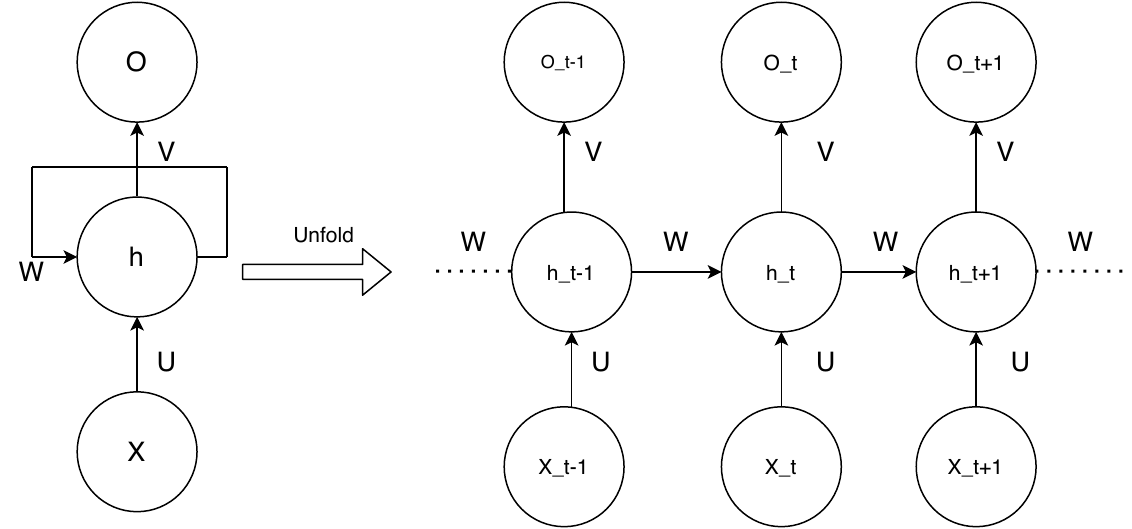} 
         \caption{RNN Architecture\cite{f2}}
         \label{f3}
     \end{subfigure}
     \\
     \begin{subfigure}[b]{0.45\textwidth}
         \centering
         \includegraphics[width=\textwidth]{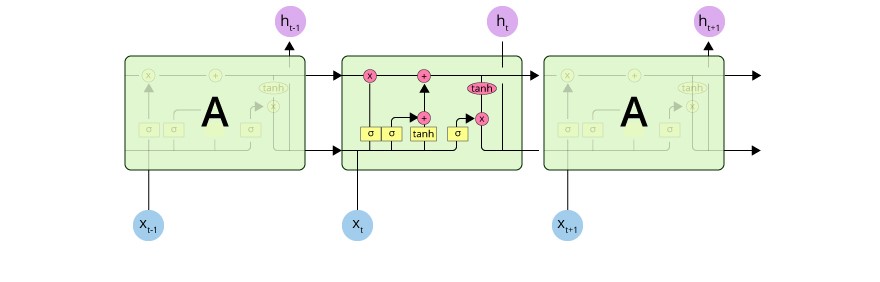}
     \caption{LSTM Architecture\cite{f2}}
         \label{f4}
     \end{subfigure}
     \hfill
     \begin{subfigure}[b]{0.45\textwidth}
         \centering
          \includegraphics[width=\textwidth]{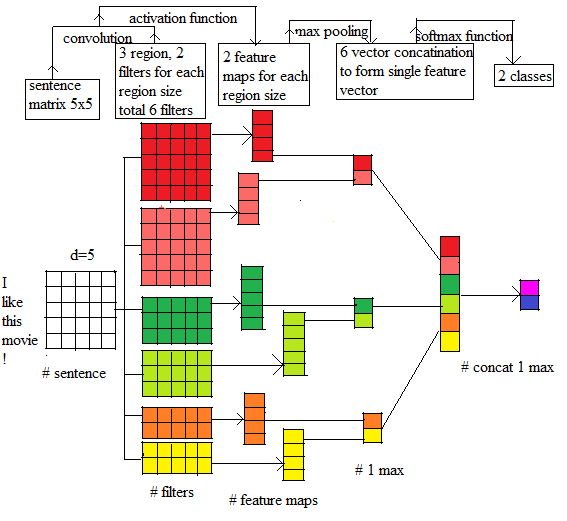}
        \caption{CNN Architecture\cite{f1}}
         \label{f5}
     \end{subfigure}
     \hfill
     \caption{Architectures of various deep learning algorithms.}
\end{figure}

In the next subsections, we briefly describe the details of the algorithms used in sentiment analysis, Table-3.
We trained our model on different N-gram models (i.e., uni-gram, bi-gram, and tri-gram models) and TP, TF, TF-IDF based weighting scheme using google apps reviews. As a result, nine different feature sets were obtained. Deep learning, a specialism of machine learning employs the adaptation of neural networks and is described in Fig. 2\cite{Y.Le} \cite{f}. The RNN, LSTM, and CNN architecture used in our study is shown in Fig.\ref{f3}, Fig.\ref{f4}, and Fig\ref{f5}.

\begin{table}[!tbp]
\centering
\caption{Algorithms Description.}
\label{t2}
\begin{tabular}{p{0.1\textwidth} p{0.85\textwidth}}
\hline
Algorithms & Summary \\
\hline
\textbf{LR}\cite{T.Hastie2009} & The idea is to come up with a model that best describes the relationship between the outcome and a set of independent variables. The dependent variable is binary, i.e., it only contains data coded as 1 (TRUE, success) or 0 (FALSE, failure). This classifier has captured a lot of work in NLP according to our literature survey, namely in teaching evaluation review\cite{Lin}, students' performance\cite{Adekitan2019}.\\

\textbf{SVM}\cite{V.Vapnik1998}& The researchers\cite{Jena2019} have worked with SVM classifiers and shown remarkable results in NLP. Teacher evaluation review also used SVM\cite{Lin}. This indeed motivated us for using SVM on our corpus too.\\

\textbf{NB}\cite{D.Lewis1998}&  It requires a small training data for classification, and all terms can be pre-computed thus, classifying becomes easy, quick, and efficient. For instance, teacher evaluation review also used NB\cite{Lin}, students' performance\cite{Adekitan2019}.\\

\textbf{KNN}\cite{D.W.Aha1991}&  It captures the idea of similarity amongst the object concerning its neighbors in terms of distance, proximity, or closeness. KNN as such did not capture much-compromising results in NLP but was used in instructors' review\cite{Onan2020}.\\

\textbf{RF}\cite{rf} & It widely used bagging, random subspace methods, the ensemble learning paradigms. For classification purpose, decision trees are used. For instance, the ensemble tree-based model\cite{Almasri2019} was efficiently modeled and showed remarkable results in NLP on students' performance dataset. Another instance was captured in students' performance\cite{Adekitan2019}.\\

\textbf{RNN} \cite{X.Li}& All the neurons are connected in a graphical form, resulting in a directed graph. The activation function used is namely, ReLu or tanh. RNN encounters vanishing gradient problem and exploding gradient problem. It cannot deal with long sequences of input. The shortcomings of RNN are easily dealt with, by using LSTMs or GRU and bidirectional RNN.
\\

\textbf{LSTM} \cite{Rojas}  &Long-term dependencies are easily handled by LSTMs and they can overcome vanishing gradient problems also. The core idea of LSTMs is to remember the information stored for long periods. They have feedback connections too. The architecture of LSTM includes an input gate, forget gate, and output gate. The flow of information in LSTM is done through cell states, by simple additions and multiplications. \\

\textbf{CNN} \cite{D.Ciresan}& Instead of general matrix multiplication, CNN uses convolution in one or more layers of the model. An input layer, an output layer and, hidden layers are included in CNN architecture. Hidden layers of CNN architecture substitute other layers namely, convolution layers, fully connected layers, normalization layers and, pooling layers. The convolution operation is employed on input data. Activation functions like ReLu, are used and add non-linearity to architecture. Pooling layers combine output from neurons and control the feature size space\cite{J.L.Elman}.
 \\ \hline
\end{tabular}
\end{table}

\subsection{Result Evaluation Metrics}

This section briefly discusses the metrics used in this study for the result computation.



\begin{equation}
F-measure=\frac{2*precision*recall}{precision+recall}
\end{equation}

\begin{equation}
accuracy=\frac{True Positive+True Negative}{True Positive+True Negative+False Positive+False Negative}
\end{equation}


\section{Research Questions}

\subsection{Research Question 1}
Do the apps which get a higher rating in the training dataset tend to be more popular among the students as well? 
\textbf{Why RQ?} Ratings help users to decide as to which app to install. According to statistics 90\% of the users consider star ratings for assessing the apps. Ratings reflect a more current version of our apps, rather than what it was years ago. The  Google Play Store Rating is re-figured and assigned more weightage to the up-to-date ratings rather than accumulating the lifetime statistics. For instance, in 2021, \textbf{Signal} overpowered \textbf{WhatsApp} in terms of ratings i.e., 4.4 and 4.1 respectively. Signal also showed a steep rise in downloads(50M+) and 1M reviews due to its Privacy amongst youngsters, and middle-aged people. 79\% of the users monitor the ratings and reviews before downloading the app; 53\% before updating apps while 55\% of them before making an in-app alteration. 0.04\% of the mass consider the ratings and reviews above suspicion rather than personalized suggestions. Apptentive indicates that \textbf{superior ratings imply more distinguished ranks\cite{rq1}. So, higher is its searching rate, with enhanced chances of being found and downloaded. No. of downloads is directly proportional to rankings.}
Complex algorithms are used to sort the search results in Google playstore based on the applicability of a search query. Ranking of apps is affected by app title, app description, in-app purchases, ratings and reviews, hidden factors, update cycle, and downloads and engagements.
Also, 15\% of the mass will contemplate downloading a 2-star rated app; 50\% of them with 3-star rated apps and 96\% of them with 4-star rated apps. The current trend in the app market suggests that 70\% of the public go through at least one of the reviews before downloading the app, 75\% say ratings inspire their app downloads. A leap from 2-star to 3-star rating eventually enhanced the app store translation by 306\%; while a leap from 3-star to 4-star rating eventually enhanced the app store translation by 96\%. In nutshell, ranking can be one of the factors that determine whether we see a given app at the top rank or bottom rank or being downloaded or not.

\subsection{Research Question 2}
 Do the priced and free apps get the same ratings and popularity from the students as compared to the training dataset? 
 
 \textbf{Why RQ?} For the developers, one of the bottlenecks of fabricating and inaugurating the mobile apps is to have minimal technical faults; making revenue out of it, is the next big hindrance. Over 90\% of apps present in app stores for free, many users are habituated to downloading the desired apps without reaching into their wallets. Apparently, app designing is not a cheap affair. Developers need to monetize to compensate for profit. Free Apps are those which are abundantly present in the play store at no monetary cost and developers monetize them through In-App Advertising. Table 4 depicts the pros and cons of paid and free apps.

\begin{table}[ht]
\centering
\caption{Paid vs Free apps.}
\begin{tabular}{p{0.1\textwidth}p{0.45\textwidth}p{0.35\textwidth}}
\hline
 & Pros & Cons \\ \hline
 Paid Apps &  
 \begin{enumerate}[nolistsep]
     \item 30 \% of the revenue is generated by the Google play store apps by charging a one-time fee from users for every download.
    \item It guarantees engagement; as the users want to make the best out of the money invested in the paid apps.
    \item The rate of competition is low in premium app categories.
    \item Paid apps are tagged with high quality, that helps generate a brand image for the organization.
 \end{enumerate}
 
 &  
 \begin{enumerate}[nolistsep]
     \item Expectation from the customers is exceptionally high as they desire outstanding customer service, valuable content, exclusive features.
    \item Number of downloads is low.
 \end{enumerate}
 \\
 Free Apps &
 \begin{enumerate}[nolistsep]
    \item Increased download quantity
    \item Lower expectations from users
 \end{enumerate}    
 & 
 \begin{enumerate}[nolistsep]
     \item No assured revenue.
     \item Lack of customer loyalty.
     \item Very high competition in the app market.
 \end{enumerate}
 \\ \hline
\end{tabular}
\label{t4}

\end{table}

\subsection{Research Question 3}
What variation in Sentiment Analysis could be analyzed by the students' reviews when compared with the training dataset? 

\textbf{Why RQ?} Sentiment analysis is done to understand the meaning and structure of a sentence. It generates customer assistance; Multimedia and Multilingual support; extract main document entities; comprehension of receipt and invoice; content classification relationship graphs; Market Research; Brand Assessment. It models not only the Polarity but at the same time, \textbf{priority}(urgent or non-urgent), \textbf{emotions}(angry, happy, sad) and, \textbf{purpose}(willing or unwilling). The challenges faced in sentiment analysis are to understand the subjectivity and tone of statements, Contexts and polarity, irony, and sarcasm; comparison of texts; emotions expressed using emojis or neutrality of statements. Sentiment Polarity is classified as Very positive(5-star), Positive, Neutral, Negative, Very negative(1-star). This helps us in real-time analysis, scaling of large datasets, and consistency. Three types of algorithms are used in sentiment analysis- 
\begin{itemize}
    \item \textbf{Rule-based} - designed using physically crafted rules.
    \item \textbf{Automatic approach} - based on machine learning paradigms.
    \item \textbf{Hybrid approach} - a mixture of rule-based and automatic approaches.
\end{itemize}

\subsection{Research Question 4}
How does the size of apps affect the installs amongst the students as compared to the training dataset? 

\textbf{Why RQ?} The app size has high repercussions on user downloads of Play Store apps. Developers have fabricated new apps due to the ever-increasing demands of the user, but one thing that concerns users is the size of the apps. For instance, according to a survey, 2.4B people are playing online games(Call of Duty- 2.7 GB and Mario Kart- 140 MB, Asphalt 8- 115 MB) on phones and this has contributed to the app's economy. Streaming services of NetFlix- 91.6 MB and Disnep+Hotstar- 30 MB  have generated \$50 M revenue. For instance, apps like, (Paytm- 44MB, Google Pay- 22MB, Pintrest- 143.1 MB) have captured the market with their advanced features and benefits. Such statistics shows that, downloading heavy-sized apps is trending amongst youth and middle-aged people. The user is bound by the smartphone’s capacity that he owns. Some of the vital issues of app size include- 
\begin{itemize}
\item\textbf{ data consumption} (some large downloads  are only supported by WiFi rather than a mobile data connection, but all users cannot afford the WiFi connectivity), 
\item\textbf{ well-ordered functionalities of the device} (sometimes due to the enormous size high-end functionalities of the app are not met due to poor data connection and lower bandwidth. Consequently, this results in uninstalls, lowers popularity amongst users), 
\item\textbf{ app downloads get slower} (based on the device's specifications, RAM, big-sized apps give warnings and pop-ups, like you need a WiFi connection to download, insufficient storage on the device and so on. This inclines the users towards smaller apps), 
\item\textbf{ phone storage gets thrashed} (sometimes the devices cannot support big apps and consume a lot of space and thereby causing heating issues, frequent switch off, hanging, other features of the device are stalled).
\end{itemize}

\subsection{Research Question 5}
What ratings are obtained for various apps based on Content Rating from students' reviews as compared to the training dataset? 

\textbf{Why RQ?} Google has laid down certain policies to govern the desired audience and app contents. For instance, \textbf{Google Play Families policies Requirements} is crafted for, apps designed for all ages above 13, apps designed for everyone or apps designed is not for children. Based on app contents the app has its privacy policies for the accessibility of apps. Sometimes Personal information also needs to be filled in the User data policy. The \textbf{Neutral age screen} authenticates the age group of the person complying with the policies. There might be unintended appeals to children. 
In the US, Google has announced an age-based rating system for its apps based on a set of policies, for instance, \textbf{Entertainment Software Rating Board (ESRB)} generates scales for the rating endorsement. Before publishing apps, developers are set to answer content-based questionnaires for their apps. ESRB works with IARC, PEGI, and USK. The new age-based rating has different categories for Everyone Teen and Mature based on certain questions' answer systems.
The statistics say that 16-24 years old are highly swayed by online reviews. They are also prone to writing negative reviews. 29\% of such age groups purchase apps even after reading critics' reviews. However, profit is earned from such young generations only and they are the ones who retaliate on app conflicts. Percentage of positive review given by age groups, 16-24=65.2\%; 25-34=79.5\%; 35-44=81.7\%; 45-54=85.6\%; 55+=85.4\%. Older groups leave a positive review as compared to the youth.66.70\% of the 16-24 age group have written an online review, the most articulated age group. 61.4\% of the 55+ age group have written an online review, the most predictive group. The market tends to be shaken by online reviews as they might seem critical. However, all age groups are taken up by online reviews.

\subsection{Research Question 6} What positive words were used by the students as compared to the training dataset? 

\textbf{Why RQ?}  Google Playstore apps have the feature to respond to user reviews. Response to the reviews helps to get in touch with users and figure out the issue. Not all reviews need to be responded to, sometimes not answering is also the best answer. There are 7 types of reviewers:
\begin{itemize}
\item \textbf{Superfans}- Such reviewers are genuine ones who like the app, put forward modifications, suggestions by narrating descriptive feedback, bugs. One needs to value such reviews and show gratitude by responding to them and grieving bonus and featuring as a top fan.
\item \textbf{Tweeters}- share their short reviews(like it, it's great!). It's not mandatory to respond to affirmative reviews as they lack importance in conversation. However negative reviews must be dealt with(a lot of errors, it's bad) as they could unfold some issues that need updating. Responses might include links to the web page or tagged with contacts.
\item \textbf{Space Cases}- includes uninformative and senseless reviews tagged with pets, celebrities, or some unpredictable events. Such irrelevant reviews needn't be responded to as it ends up in a debate with strangers in comments.
\item \textbf{Haters}- they tend to find flaws in every app and remain unhappy always and sharing filthy reviews. Developers and users would like to reciprocate them but people should be dealt with them wisely as all the comments are public. They might be depicting a valid judgment in an unfriendly manner. It’s unnecessary to respond to their silly comments and 
\item \textbf{Browsers} - they try the latest app weekly by downloading numerous apps and share their instantaneous reviews without using the app for an hour. They are sometimes annoying but sometimes give valuable judgment by comparing different apps. They aren't true-hearted users.
 \item \textbf{Spammers} - for the sake of promotions the spammers might  create futuristic reviews. It's best to avoid them and flag list them spam.
\item \textbf{Techies} - include developers or a person with great insight who knows much about the app functionalities  and provide intuitive judgments. They are similar to Superfans and must be taken care of by tagging them to the web page or linking to their contact. Thanking them for their feedback is necessary.
\end{itemize}
So, based on the aforesaid classification of reviewers, we may infer, Superfans, Tweeters, and Techies are the ones who share their positive feedback over social media.

\subsection{Research Question 7}
What negative words were used by the students as compared to the training dataset?

\textbf{Why RQ?}Based on the above classification in Section 4.6, we may say, Spacecases, Spammers, Browsers, and Haters are the ones who share their negative feedback over social media. Reviews and ratings come in many forms, and some are more valuable than others. It would be nice to think that the overall rating or score of an app on the App Store or Google Play is a reliable indicator of an app's actual quality. The reviews and ratings for the app from different users have discrete practicality in the app market. Apps' actual grading is highly dependent on the overall rating. However, the bottleneck is that all reviews are not alike. Reviews are categorized into: 
\begin{itemize}
\item\textbf{One-word wonder} - it includes stereotyped reviews having no tags with ratings rather all are of the same length(example, Fantastic, awesome). These don't contribute to constructive criticism. These are of no importance to build customer or app links. Such reviews must be featured or marked as helpful by others to show their authenticity.   
\item \textbf{Hidden treasure}- it includes lengthy reviews by the users who donate a lot of time in writing because they are either bothered or furious over the apps. They narrate their experience, share feedback describing proposals or circumstances. Such reviews should be taken into consideration and appreciating them could be valuable.
\item \textbf{Subjective Responder}- it is similar to hidden treasure. Such reviewers are highly judgemental because they might have faced some hassle, thereby giving numerous feedback and low ratings(1 or 2-star). Hence, they are biasing their reviews. Such reviews are hard to differentiate between their subjectivity and criticism. So, the permissible flaws of the apps are sometimes overlooked.
\item \textbf{One-star until…} - reviewers describe their perspectives when they face complications in their apps and rate low scores to apps to woo the developers. Such negative reviews might be demotivating. Reviewers may fail to delete their reviews when their issues are sorted out. This might be deceptive or sometimes reveal the authenticity(positive criticism) of the apps giving the potential users a general vision about the apps. Such reviews may help the developers to rectify their flaws by fixing out the complaints and improve the app's rating in near future. 
\item \textbf{Spammers} - the app market is filled with top-earning companies and app developers with high revenues, but they are prone to leaving contradictory feedback about other apps. Google Play pays attention to such spam by reporting them. But sometimes it may affect the app conversion rate.
\end{itemize}
From the above classification, we may say, the subjectivity of aforesaid reviews fall under negative criticism.

\subsection{Research Question 8}
What is the correlation between price, rating, popularity amongst the university students when compared with the training dataset? 

\textbf{Why RQ?} The pricing strategy of apps comprises of- \begin{enumerate}
    \item \textbf{Paid}(pay once for every download),
    \item \textbf{Free}(easily available without money)
    \item \textbf{trial}(pay for a span of 7 or 15 days)
    \item \textbf{Freemium}(freely available with limited features and functionalities)
    \item \textbf{Paidmium}(aggregation of Paid and Freemium)
\end{enumerate}
Paid apps get a high rating and popularity if the value of the product is always more than the price;  designed as per user specifications; meets the high expectations of users; comprehends the market requirements and a rational amount is charged. The likeliness to download Free apps is high so the popularity and ratings of such apps are also high. The download cycle is dependent on the App Store Optimization (ASO) Loop, which in turn gives higher visibility to new users.

\subsection{Research Question 9} How close are students' reviews as compared to the training dataset?

\textbf{Why RQ?}  We may use Confusion matrix to answer this. The usefulness of the Confusion matrix is that it indicates the model’s confusion in making predictions. It gives intuitions about the errors and its types the classifier makes. The columns of the matrix representation in the confusion matrix illustrate the predicted class specimen while the rows depict the actual class specimen. Computing the accuracy and f-score highlights the errors well. Our study deals with multinomial classification problems, having 3 variates(Positive, Negative, and Neutral sentiment) so the Confusion matrix and class statistics can be easily expanded.

\subsection{Research Question 10} How close are students' reviews as compared to the training dataset on application of GloVe scheme?

\textbf{Why RQ?}  The visualization aspect of the confusion matrix persuaded us to use it in our study to make predictive judgments. One of the motivations to use confusion matrix with the application of GloVe scheme was to compute the model's functionalities for the true values based on the misclassifications. This implementation i.e., Confusion matrix with GloVe,  is useful when we have imbalanced dataset (i.e., dataset having non-uniform distribution of class labels.). In our study, we had a real-time dataset and to deal with real-life problems we have imbalanced class distribution. This implementation was done to highlight how crucial False Negatives(FN) and False Positives(FP) were for our study. A low FN and FP values implied correct identification of real sentiments and not disturbed by false sentiments. 


\section{Experiments and Results}


The language used is Python and the platform used was Jupyter Notebook. 
In this study, we have leveraged the power of Google Colab to develop our machine learning and deep learning-based sentiment classifier. Furthermore, the code for the experimental analysis is available at  "https://github.com/smlab-niser/Google-Reviews-Sentiment-Analysis".

\begin{figure}
     \centering
     \begin{subfigure}[b]{1\textwidth}
         \centering
             \includegraphics[width=\textwidth]{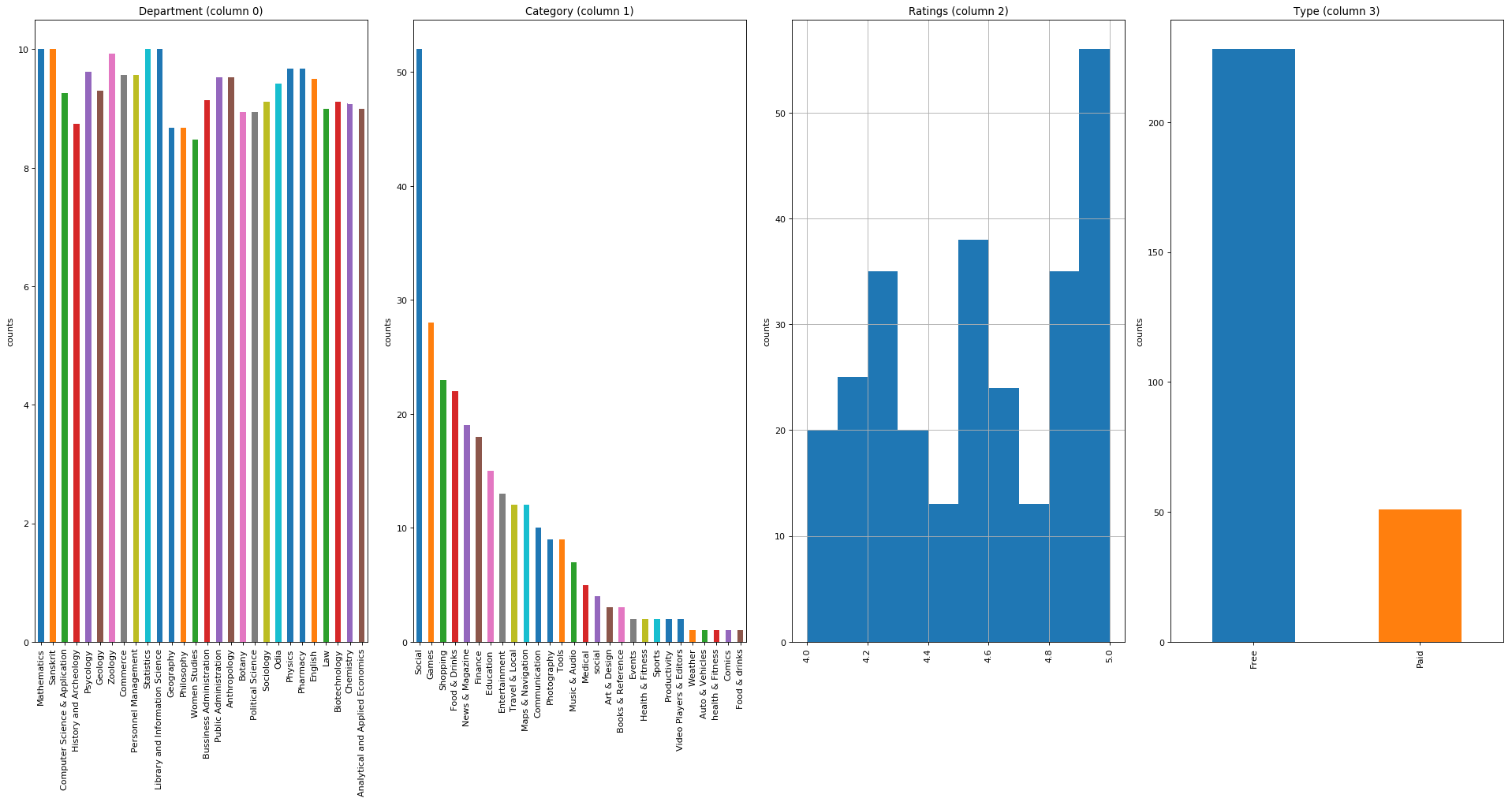}
            \caption{Distribution of Counts, in terms of no. of app downloads by university students over apps' Category, Ratings, Departments, and, Price strategy from the Playstore using training and test dataset.}
            \label{rq1-1}
     \end{subfigure}
     \hfill \\
     \begin{subfigure}[b]{0.7\textwidth}
         \centering
          \includegraphics[width=\textwidth]{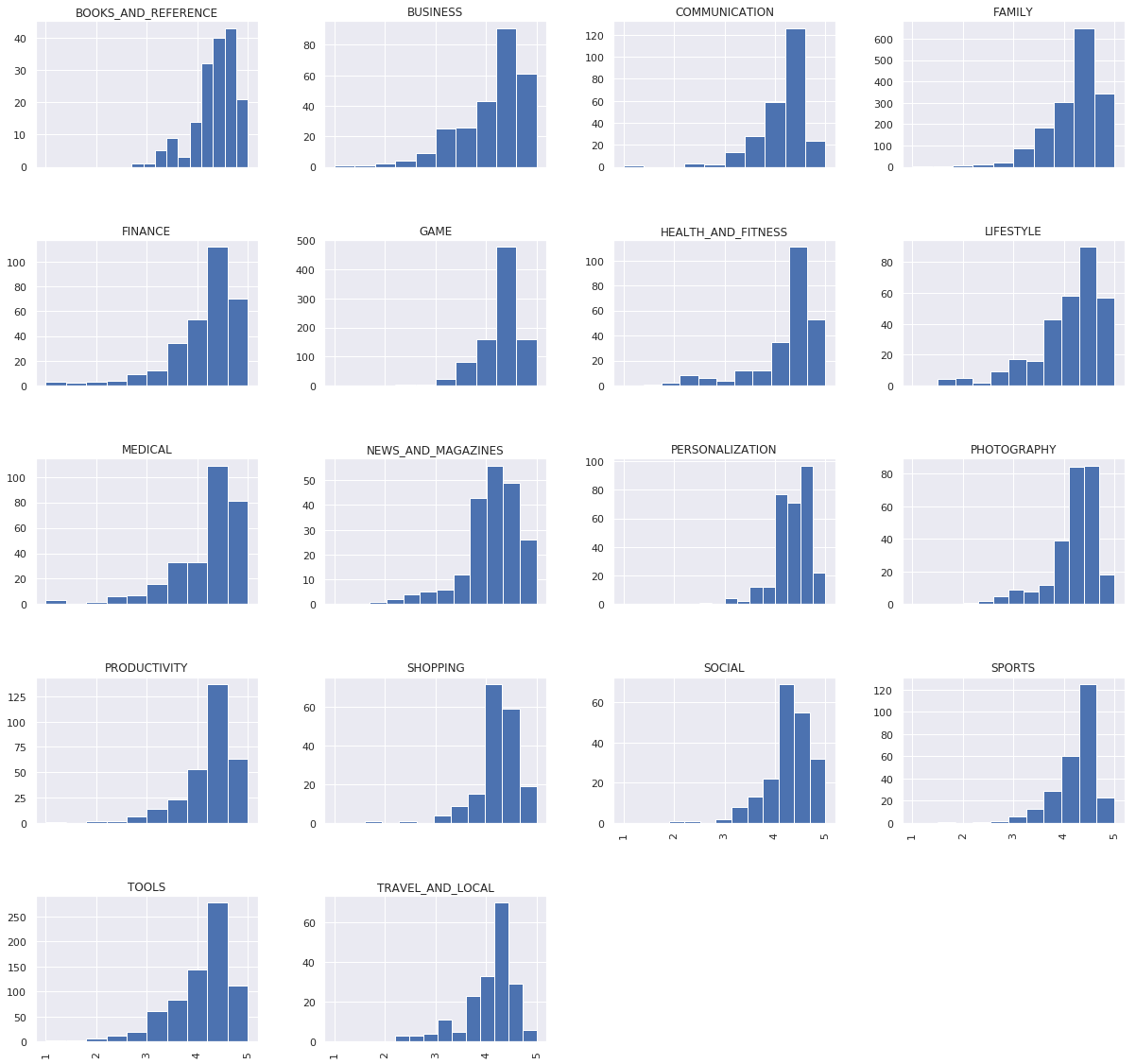}
    \caption{ Distribution of apps' Ratings over apps' Category of Playstore on the basis of download strategy by university students using training and test dataset.}
      \label{rq1-2}
     \end{subfigure}
     \hfill
     \caption{Results for RQ1}
\end{figure}

\begin{figure}
     \centering
     \begin{subfigure}[b]{0.45\textwidth}
         \centering
             \includegraphics[width=\textwidth]{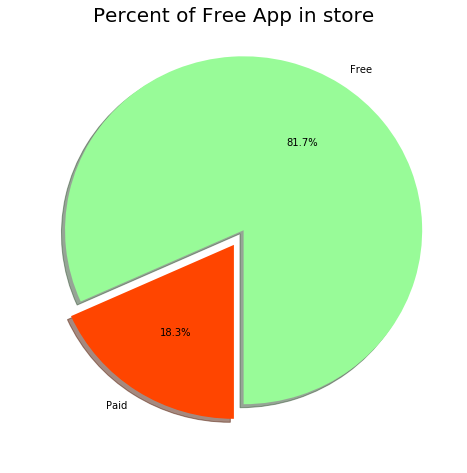}
            
            \caption{Distribution of Paid apps vs Free apps of Playstore based on the download strategy by the university students using training and test dataset.}
            \label{rq2-1}
     \end{subfigure}
     \hfill 
     \begin{subfigure}[b]{0.45\textwidth}
         \centering
          \includegraphics[width=\textwidth]{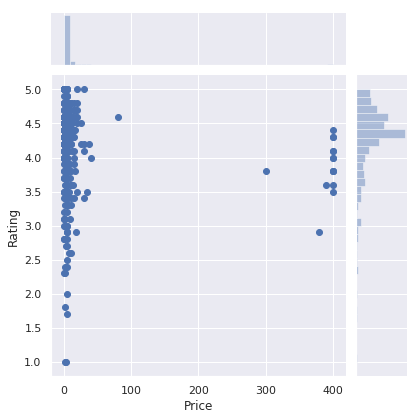}
       
         \caption{Distribution of Priced apps of Playstore vs app Ratings based on the download strategy by the university students using training and test dataset.}
          \label{rq2-2}
     \end{subfigure}
     \hfill
     \caption{Results for RQ2}
\end{figure}
 
\begin{figure}
    \centering
    \includegraphics[width=0.5\textwidth]{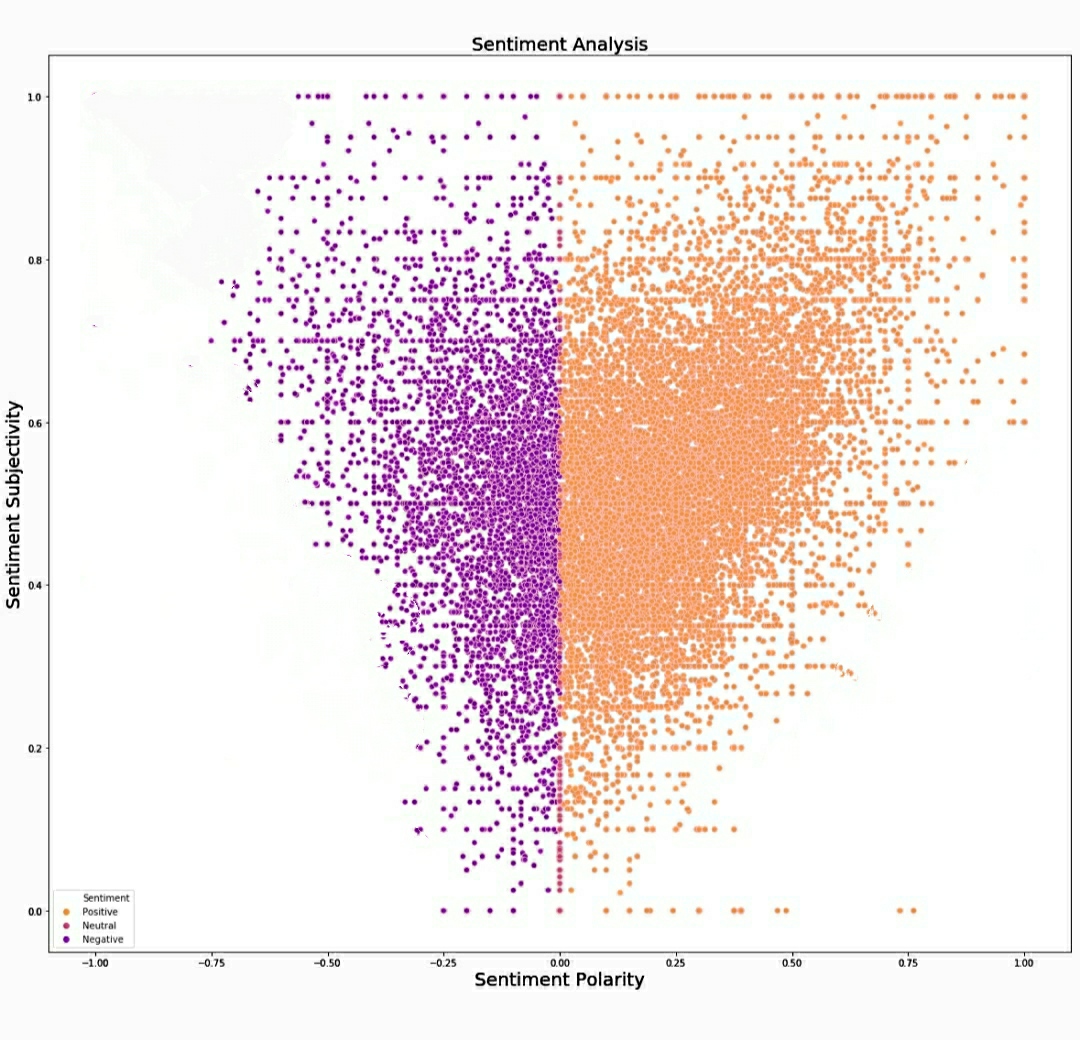}
    \caption{RQ3 - Distribution of Sentiment Subjectivity over Sentiment Polarity to analyze the sentiments of university students towards the apps of Playstore using training and test dataset.}
         \label{rq3}
\end{figure}
 
\begin{figure}
 \begin{minipage}[b]{0.3\textwidth}
    \includegraphics[width=\textwidth]{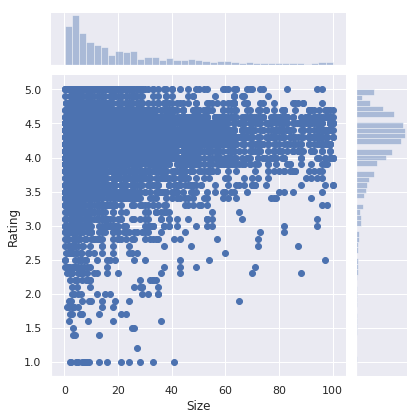}
      \caption{RQ4 - Distribution of Ratings of the apps of Playstore vs their Sizes based on the download strategy by the university students and training dataset.}
        \label{rq4}
  \end{minipage}
  \hfill
   \begin{minipage}[b]{0.6\textwidth}
      \includegraphics[width=\textwidth]{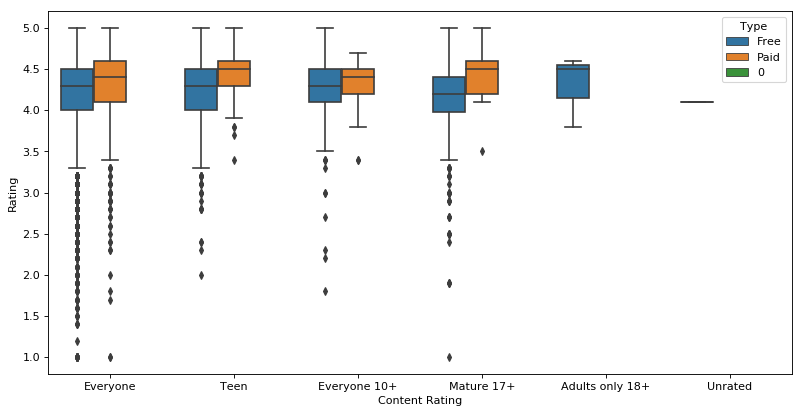}
        \caption{RQ5 - Distribution of Ratings of the apps of Playstore vs their Content Ratings based on the download strategy by the university students and training dataset.}
         \label{rq5}
    \end{minipage}
    \end{figure}

\subsection{Results based on Visualization}
 One of the motivations to create RQ's was to analyze the personal skill of students and showcase the fundamentals of statistics merged with programming. 
 Whatever we speak, listen or write is in the form of natural language(namely, WhatsApp chat messages, movie dialogues, etc.). 
So, we set out to answer the questions in context with basic characteristics of the app market like price, category, ratings, genres, size, and downloads. 
The results obtained were compared with the SAR through visualizations described below in Table-\ref{t5}. The inferences drawn using charts helped us to get a glimpse of students' behavior towards the distribution of the app market. 
\newline



\begin{longtable}[c]{p{7em} p{7em} p{30em}}
\caption{Assessing the Research Questions.}
\label{t5}\\
\hline
Research Question & Figure & Answer \\
\hline
\endhead
 RQ1 &  Fig.\ref{rq1-1},\ref{rq1-2}. &  Section 4.1 emphasizes on a general note whether the apps which get a higher rating in the training dataset tend to be more popular among the mass or not. The current trend in the app market suggests that 70\% of the public go through at least one of the reviews before downloading the app, 75\% say ratings inspire their app downloads. However, from the visualization plot, we got a glimpse of the students’ behavior and the inference drawn was, the Google app market breakdown showed prominent downloads in Social and Games categories. On the contrary, Weather and Comics were of least interest among students. The average ratings shooted to 4.17 across major categories. Interestingly, Shopping, Food, and Drinks, News, and Magazine are also catching up. Expensive apps may make students disappointed if they are not good enough and consequently, get low ratings. Students from the Mathematics and Sanskrit department participated fairly well while Women Studies and Geography showed the least participation. Other departments showed acceptable participation.  \\
  
RQ2 &  Fig.\ref{rq2-1},\ref{rq2-2}. & Section 4.2 emphasizes on a general note whether the priced and free apps get the same ratings and popularity from the public or not. The developers' aim is to generate revenues from the apps. However, from the visualization plot, we got a glimpse of the students’ behavior and the inference drawn was, the percentage of free apps(81.7\%) exceeds the paid apps(18.3\%) in terms of download strategy. While jointplot visualization depicts the pricing strategy where the points are heavily clustered for unpaid apps. This gives us an inference that students prefer free apps rather than paid and an average rating between 3.5 to 5 is shown. \\

RQ3  &    Fig.\ref{rq3}  &  Section 4.3 emphasizes on a general note the variation of the sentiments based on subjectivity and polarity when assessed for the public based on the training dataset. The priority(urgent or non-urgent), emotions(angry, happy, sad) and, purpose(willing or unwilling) of the text are showcased through sentiment analysis. However, from the visualization plot, we got a glimpse of the students’ behavior and the inference drawn was, in the scatter plot between sentiment polarity and subjectivity the points are heavily clustered towards the positive side rather than on the negative. Hence, covering a major area towards the right side between 0 to 1 on the x-axis. Specifically, we can say students weren't so harsh while giving reviews, instead gave genuine and lenient feedback.\\ 
 
RQ4 &  Fig.\ref{rq4}.   & Section 4.4 emphasizes on a general note for the installs of the apps based  on the size of apps based on the training dataset. We found app size is directly proportional to public demands and the app’s economy. Large apps are directly proportional to downloads amongst youth and middle-aged people.
However, from the visualization plot, we got a glimpse of the students’ behavior and the reverse trend was observed. The inference drawn was, the jointplot depicts the sizing strategy(small vs huge). We got a clear conclusion that small-sized apps(0-60 Mb)  are predominant for downloads among students. This enhances the ratings. The average rating turned out to be 4-5. On the contrary, larger apps have fewer ratings and are less preferable.  
This might be because students have different categories of devices supporting different storage and data was gathered from a very limited domain of students itself.\\

RQ5  &      Fig.\ref{rq5}.    &  Section 4.5 emphasizes on a general note the variation in ratings amongst the public for various Content ratings where the statistics reveal that 16-24 years old are highly influenced by online reviews and write negative reviews. Percentage of positive review given by age groups, 16-24=65.2\%; 25-34=79.5\%; 35-44=81.7\%; 45-54=85.6\%; 55+=85.4\%. However, from the visualization plot, we got a glimpse of the students’ behavior and the inference drawn was, the free and paid apps showed a predominance in ratings i.e., 4-4.5 amongst students that were accessible to Everyone. Free apps that were accessible to Adults only 18+ showed a rating of 4.5. The free and paid apps under the section Teens, Everyone 10+, Mature 17+ showed an average rating of 3.2-4.8. \\

 RQ6    &    Fig.\ref{rq6}.   & Section 4.6 emphasizes on a general note the types of reviewers based on the general public and their school of thought for giving a positive reviews. From amongst the masses, Superfans, Tweeters, and, Techies are the ones who share their positive feedback over social media. However, from the visualization plot, we got a glimpse of the students’ behavior and the inference drawn was, the bold and highlighted words(good, great, love)were highly used amongst students, while smaller and less distinct words(little, much, back) were least used.                       \\

 RQ7 &  Fig.\ref{rq7}. &   Section 4.7 emphasizes on a general note the types of reviewers based on the general public and their school of thought for giving a negative reviews. From amongst the masses,  Space Cases, Spammers, Browsers, and Haters are the ones who share their negative feedback over social media. However, from the visualization plot, we got a glimpse of the students’ behavior and the inference drawn was, the bold and highlighted words(load, log, work, take-time) were highly used amongst students, while smaller and less distinct words(problem, open, login) were least used.\\

 RQ8 & Fig.\ref{rq8}. & Section 4.8 emphasizes on a general note the correlation between price, rating, popularity amongst the public based on the training dataset. Free apps are found to be common amongst the masses, while paid apps become flexible to download when user specifications and requirements are met. However, from the visualization plot, we got a glimpse of the students’ behavior and the inference drawn was, the installs and reviews are positively correlated amongst students. While installs and pricing is negatively correlated. \\
 
 RQ9 &    Fig.\ref{rq9}.   &  Section 4.9 emphasizes on a general note the similarity in reviews amongst the public. Confusion matrix is the best tool to make intuitions regarding the misclassifications and errors a classifier makes. However, from the visualization plot, we got a glimpse of the students’ behavior and the inference drawn was, the confusion matrix when applied with LR gave us an accuracy of 90.8\%.              \\ 
 
 RQ10  &  Fig.\ref{rq10}.  & Section 4.10 emphasizes on a general note the similarity in reviews amongst the public using GloVe technique. This helped us to judge the model’s performance for true values based on misclassifications. This technique helped us a lot in the imbalance dataset. However, from the visualization plot, we got a glimpse of the students’ behavior and the inference drawn was, the confusion matrix when applied with GloVe gave us the highest accuracy of 81.6 on Falsely classified reviews and second-highest accuracy of 80.7\% on truly classified reviews.                   \\ 
 
\hline
\end{longtable}


\begin{figure}[!tbp]
  \centering   
  \begin{minipage}[b]{0.4\textwidth}
    \includegraphics[width=\textwidth]{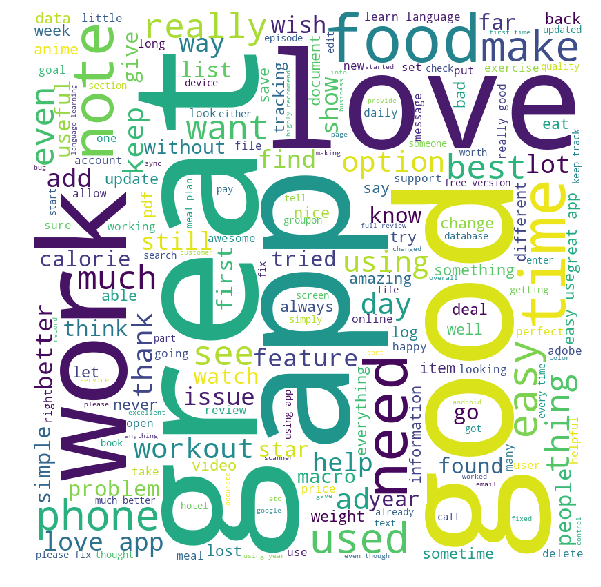}
     \caption{RQ6 - Distribution of all the Positive Words in the Word Cloud based on the University students' reviews and training dataset for the Playstore apps.}
      \label{rq6}
  \end{minipage}
  \hfill
  \begin{minipage}[b]{0.4\textwidth}
    \includegraphics[width=\textwidth]{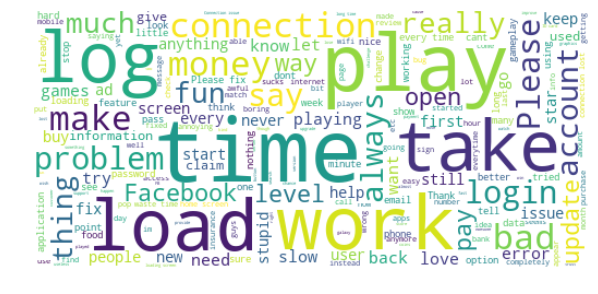}
      \caption{RQ7 - Distribution of all the Negative Words in the Word Cloud based on the University students' reviews and training dataset for the Playstore apps.}
         \label{rq7}
  \end{minipage}

\end{figure}

\begin{figure}
    \centering
      \begin{minipage}[b]{0.7\textwidth}
        \includegraphics[width=\textwidth]{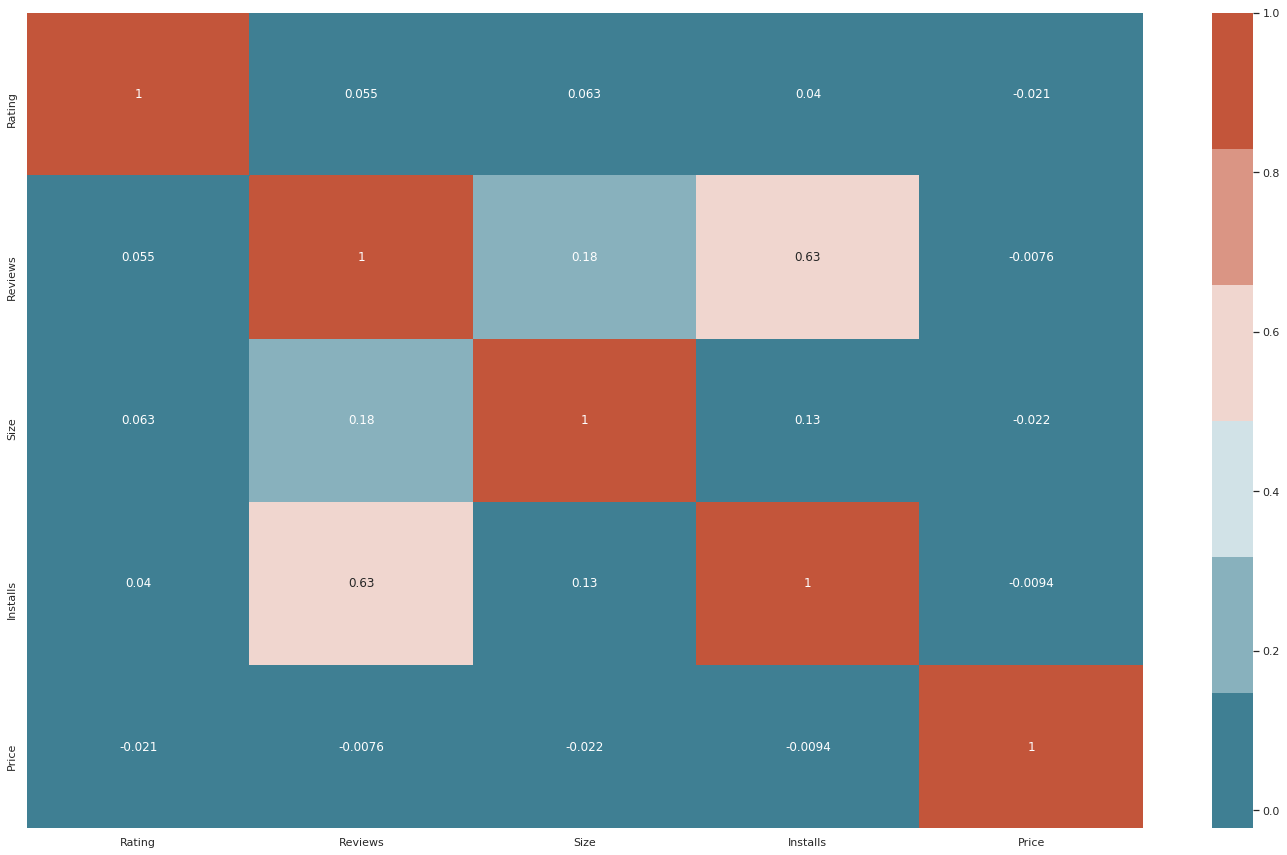}
         \caption{RQ8 - Heatmap depicting the Correlation between Playstore apps' features - Price, Installs, Size, Reviews and, Ratings using the training and test data set.}
         \label{rq8}
      \end{minipage}
  \end{figure}
  
\begin{figure}
     \centering
      \begin{minipage}[b]{0.3\textwidth}
         \centering
         \includegraphics[width=\textwidth]{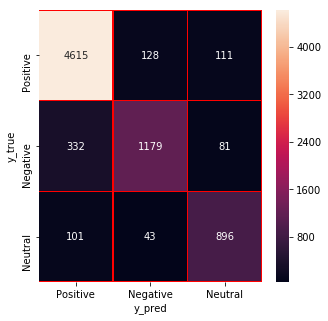}
         \caption{RQ9 - Confusion Matrix depicting the performance of the classifier based on the university students' reviews and training dataset.}
         \label{rq9}
     \end{minipage}
     \hfill
      \begin{minipage}[b]{0.3\textwidth}
         \centering
         \includegraphics[width=\textwidth]{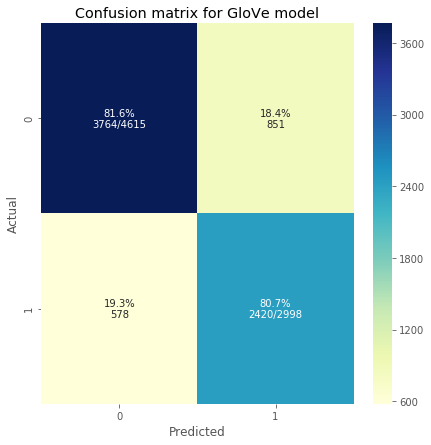}
         \caption{RQ10 - Confusion Matrix with Glove strategy depicting the performance of the classifier based on the university students' reviews and training dataset.}
         \label{rq10}
     \end{minipage}
     \hfill
    
    \end{figure}

\subsection{Results based on Machine Learning}

Secondly, we performed nine experiments to train our model using the classification algorithms( NB, LR, KNN, SVM, and RF) on the conventional text representation schemes(TP, TF, TF-IDF).
The ensemble learning method namely Bagging was also used in our empirical analysis. 
We used evaluation metrics namely F-score and accuracy to generate useful intuitions from our corpus; Table-\ref{t6}, Table-\ref{t7}. 
We infer that SVM proved best for our corpus and attained the highest accuracy and F-score value. SVM on bi-gram+TF-IDF got accuracy 93.41\% and TF-IDF on bi-gram, tri-gram  model got an F-score 0.89. NB performed worst on our corpus and didn't turn out to be fit for our study. The least accuracy of 78.56\% and an F-score of 0.60 was achieved. The second best algorithm of our study is KNN and is catching up with SVM. TF on bi-gram got the highest accuracy of 91.5\% and an F-score of 0.85 was constant throughout. LR and RF performed averagely on our corpus in terms of F-score and accuracy. LR on uni-gram and TF schemes resulted in an accuracy of 84.99\%, F-score of 0.70. RF on bi-gram+TP captured the highest accuracy 85.47\%., F-score 0.68 on uni-gram models for all TP, TF, TF-IDF schemes. Bagging was applied on LR and NB and we got comprehensible results showing an enhancement in accuracy and F-score. 
\begin{table}[]
\centering
\caption{Accuracy values for Machine Learning Algorithms.}
\label{t6}
\begin{tabular}{llllllll}
\hline
 & SVM & KNN & LR & RF & NB & \begin{tabular}[c]{@{}l@{}}LR\\ (Bagging)\end{tabular} & \begin{tabular}[c]{@{}l@{}}NB\\ (Bagging)\end{tabular} \\ \hline
\begin{tabular}[c]{@{}l@{}}Unigram+\\ TP\end{tabular} & 91.5 & 90 & 84.36 & 83 & 78.56 & 86.47 & 85.14 \\
\begin{tabular}[c]{@{}l@{}}Unigram+\\ TF\end{tabular} & 92 & 91 & 84.99 & 84.15 & 79.25 & 86.5 & 85.69 \\
\begin{tabular}[c]{@{}l@{}}Unigram+\\ TF-IDF\end{tabular} & 92.89 & 91.01 & 84.08 & 83.42 & 80 & 86.5 & 85.5 \\
\begin{tabular}[c]{@{}l@{}}Bigram+\\ TP\end{tabular} & 93.4 & 91 & 84.77 & 85.47 & 81.2 & 86.5 & 85 \\
\begin{tabular}[c]{@{}l@{}}Bigram+\\ TF\end{tabular} & 93 & 91.5 & 84.96 & 84.23 & 82.09 & 86.77 & 86.69 \\
\begin{tabular}[c]{@{}l@{}}Bigram+\\ TF-IDF\end{tabular} & 93.41 & 90.9 & 84.61 & 85.11 & 82.14 & 86.5 & 85.11 \\
\begin{tabular}[c]{@{}l@{}}Trigram+\\ TP\end{tabular} & 93 & 89.5 & 85 & 84.5 & 81.27 & 86.5 & 85.68 \\
\begin{tabular}[c]{@{}l@{}}Trigram+\\ TF\end{tabular} & 93 & 89 & 84 & 85 & 80 & 87 & 85 \\
\begin{tabular}[c]{@{}l@{}}Trigram+\\ TF-IDF\end{tabular} & 93.37 & 88.39 & 84.48 & 84.16 & 82.21 & 87.88 & 84 \\ \hline
\end{tabular}
\end{table}

\begin{table}[]
\centering
\caption{F score values for Machine Learning Algorithms.}
\label{t7}
\begin{tabular}{llllllll}
\hline
 & SVM & KNN & LR & RF & NB & \begin{tabular}[c]{@{}l@{}}LR\\ (Bagging)\end{tabular} & \begin{tabular}[c]{@{}l@{}}NB\\ (Bagging)\end{tabular} \\ \hline
\begin{tabular}[c]{@{}l@{}}Unigram+\\ TP\end{tabular} & 0.88 & 0.85 & 0.7 & 0.68 & 0.7 & 0.85 & 0.77 \\
\begin{tabular}[c]{@{}l@{}}Unigram+\\ TF\end{tabular} & 0.88 & 0.85 & 0.7 & 0.68 & 0.7 & 0.85 & 0.78 \\
\begin{tabular}[c]{@{}l@{}}Unigram+\\ TF-IDF\end{tabular} & 0.89 & 0.86 & 0.69 & 0.68 & 0.72 & 0.87 & 0.75 \\
\begin{tabular}[c]{@{}l@{}}Bigram+\\ TP\end{tabular} & 0.87 & 0.85 & 0.66 & 0.63 & 0.62 & 0.87 & 0.75 \\
\begin{tabular}[c]{@{}l@{}}Bigram+\\ TF\end{tabular} & 0.87 & 0.85 & 0.66 & 0.6 & 0.6 & 0.86 & 0.75 \\
\begin{tabular}[c]{@{}l@{}}Bigram+\\ TF-IDF\end{tabular} & 0.89 & 0.85 & 0.68 & 0.61 & 0.62 & 0.86 & 0.76 \\
\begin{tabular}[c]{@{}l@{}}Trigram+\\ TP\end{tabular} & 0.88 & 0.85 & 0.71 & 0.62 & 0.64 & 0.84 & 0.77 \\
\begin{tabular}[c]{@{}l@{}}Trigram+\\ TF\end{tabular} & 0.88 & 0.84 & 0.7 & 0.63 & 0.63 & 0.85 & 0.77 \\
\begin{tabular}[c]{@{}l@{}}Trigram+\\ TF-IDF\end{tabular} & 0.88 & 0.85 & 0.7 & 0.62 & 0.63 & 0.86 & 0.76 \\ \hline
\end{tabular}
\end{table}

\subsection{Results based on Deep Learning}
Furthermore, we performed experiments to train our model using the deep learning algorithms also using word embedding, namely GloVe. 
The Google app reviews were evaluated using the major deep learning algorithms namely, LSTM, RNN, and CNN briefly described in Section 3.6. 
The batch size and hyper-parameters were thoroughly investigated during the experiments. The vector size used in our experiment was 200 while the dimension of the projection layer was 100. The different configurations so obtained, shows a comparison amongst these deep algorithms based on accuracy and F-scores. 
Regarding the performance of algorithms listed in Table-\ref{t8}, LSTM with GloVe attained the highest accuracy of 95.2\%, and an F-score of 0.88. CNN and RNN with GloVe performed averagely with an accuracy value close to 93\% 
In terms of F-score, CNN with Glove proved worst for our corpus by attaining 0.78.

\begin{table}[ht]
\centering
\caption{Accuracy and F-score values for Deep Learning Algorithms. }
\label{t8}
\begin{tabular}{lllll}
 Algorithms & Vector Size & Dimension of Projection layer &Accuracy & F-score \\
\hline
LSTM(GloVe) & 200 & 100 & 95.2 & 0.88 \\
RNN(GloVe) & 200 & 100 & 93 & 0.85 \\
CNN(GloVe) & 200 & 100  & 92.7 & 0.78\\
\hline
\end{tabular}
\end{table}

\subsection{Main Effects Plots for Empirical Analysis}

\begin{figure}[!tbp]
  \centering
 \begin{minipage}[b]{0.45\textwidth}
       \includegraphics[width=\textwidth]{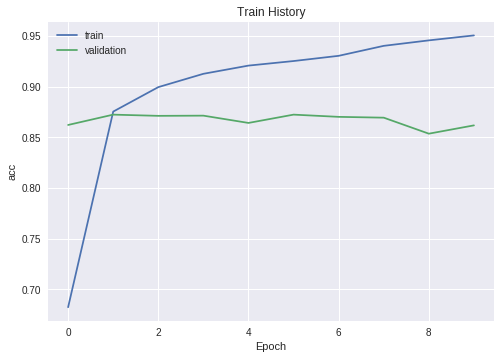}
         \subcaption{ Accuracy Plot for training data for LSTM.}
         \label{f18}
   \end{minipage}
   \hfill
   \begin{minipage}[b]{0.45\textwidth}
     \includegraphics[width=\textwidth]{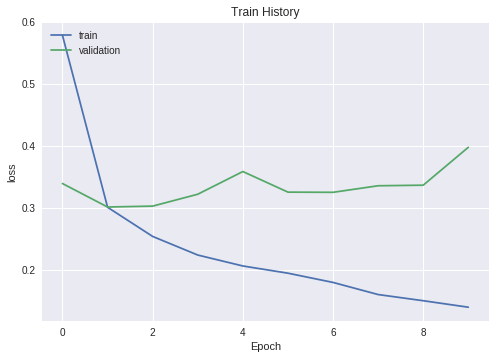}
       \subcaption{  Loss Plot for training data for LSTM.}
         \label{f19}
  \end{minipage}
  \hfill
  \begin{minipage}[b]{0.45\textwidth}
    \includegraphics[width=\textwidth]{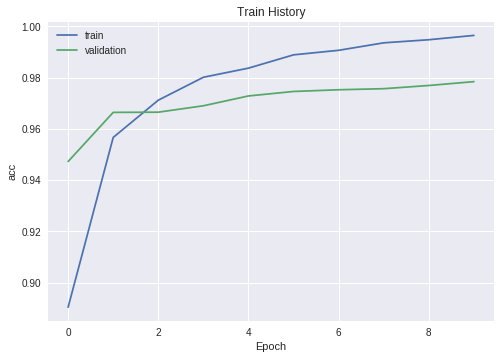}
      \subcaption{ Accuracy Plot for test data for LSTM.}
         \label{f20}
  \end{minipage}
  \hfill
  \begin{minipage}[b]{0.45\textwidth}
    \includegraphics[width=\textwidth]{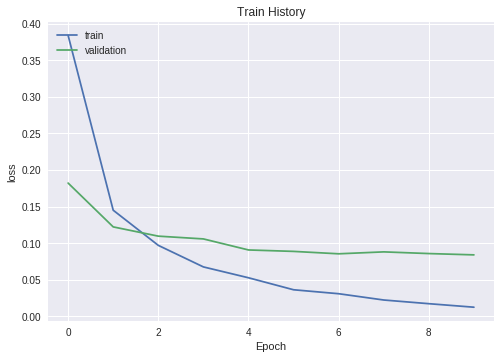}
       \subcaption{ Loss Plot for test data for LSTM.}
         \label{f21}
  \end{minipage}
  \caption{Comparative Plots for LSTM.}
\end{figure}

\begin{figure}[!tbp]
  \centering
\begin{minipage}[b]{0.45\textwidth}
      \includegraphics[width=\textwidth]{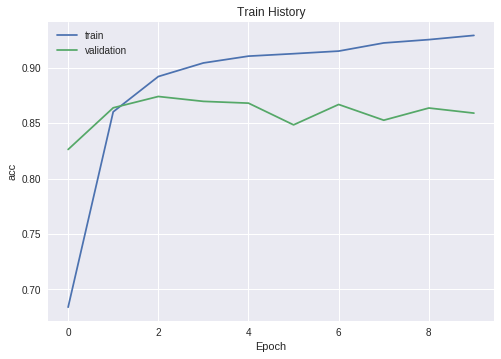}
        \subcaption{  Accuracy Plot for training data for RNN.}
          \label{f22}
  \end{minipage}
  \hfill
  \begin{minipage}[b]{0.45\textwidth}
    \includegraphics[width=\textwidth]{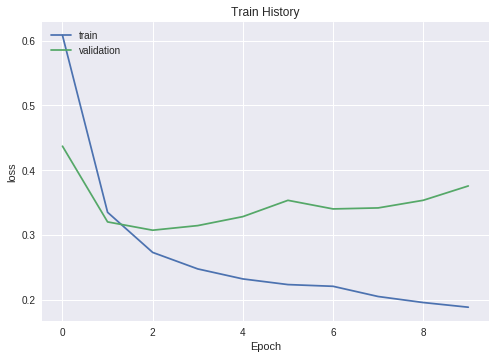}
      \subcaption{ Loss Plot for training data for RNN.}
         \label{f23}
  \end{minipage}
  \hfill
   \begin{minipage}[b]{0.45\textwidth}
    \includegraphics[width=\textwidth]{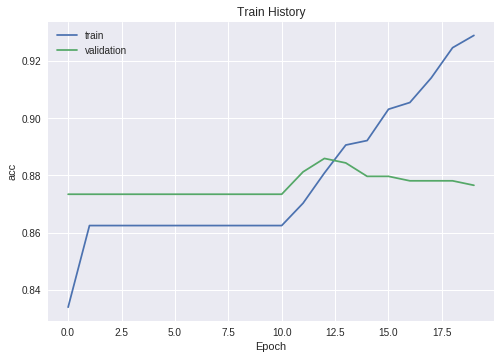}
     \subcaption{ Accuracy Plot for test data for RNN.}
        \label{f24}
  \end{minipage}
\hfill
  \begin{minipage}[b]{0.45\textwidth}
    \includegraphics[width=\textwidth]{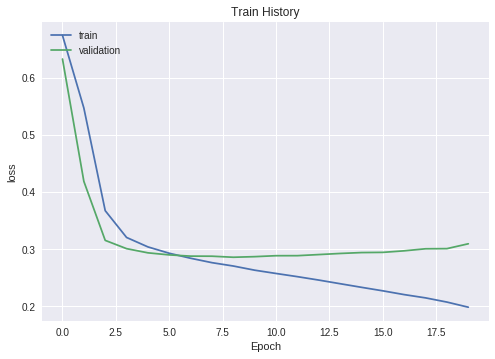}
      \subcaption{ Loss Plot for test data for RNN.}
        \label{f25}
  \end{minipage}
    \hfill
  \caption{Comparative Plots for RNN.}
\end{figure}

\begin{figure}[!tbp]
  \centering
  \begin{minipage}[b]{0.45\textwidth}
     \includegraphics[width=\textwidth]{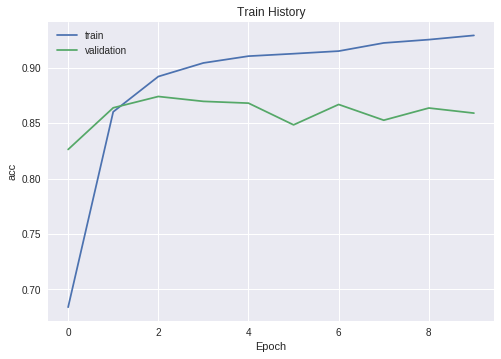}
       \subcaption{Accuracy Plot for training data for CNN.}
         \label{f26}
  \end{minipage}
  \hfill
  \begin{minipage}[b]{0.45\textwidth}
    \includegraphics[width=\textwidth]{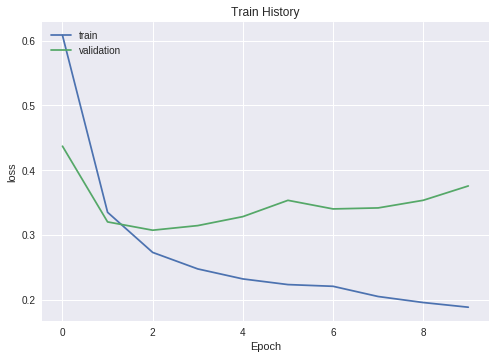}
      \subcaption{Loss Plot for training data for CNN.}
        \label{f27}
  \end{minipage}
  \hfill
  \begin{minipage}[b]{0.45\textwidth}
    \includegraphics[width=\textwidth]{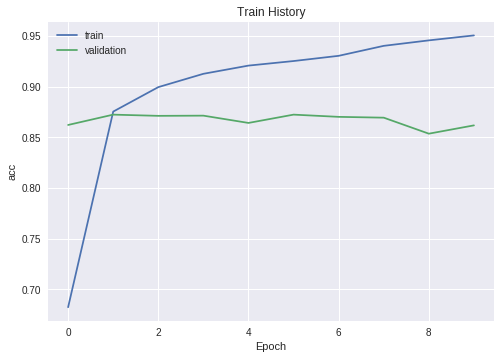}
      \subcaption{  Accuracy Plot for test data for CNN.}
         \label{f28}
  \end{minipage}
  \hfill
  \begin{minipage}[b]{0.45\textwidth}
    \includegraphics[width=\textwidth]{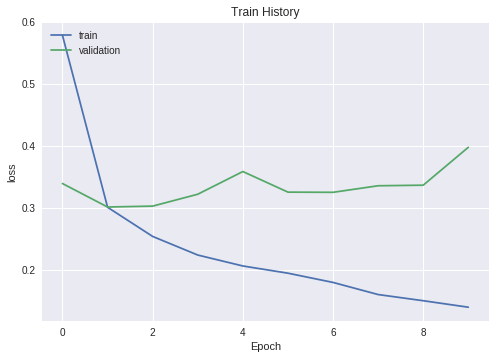}
        \subcaption{Loss Plot for test data for CNN.}
          \label{f29}
  \end{minipage}
  \caption{Comparative Plots for CNN.}
  \hfill
\end{figure}

\begin{figure}[!tbp]
    \centering
 \begin{minipage}[b]{0.4\textwidth}
       \includegraphics[width=\textwidth]{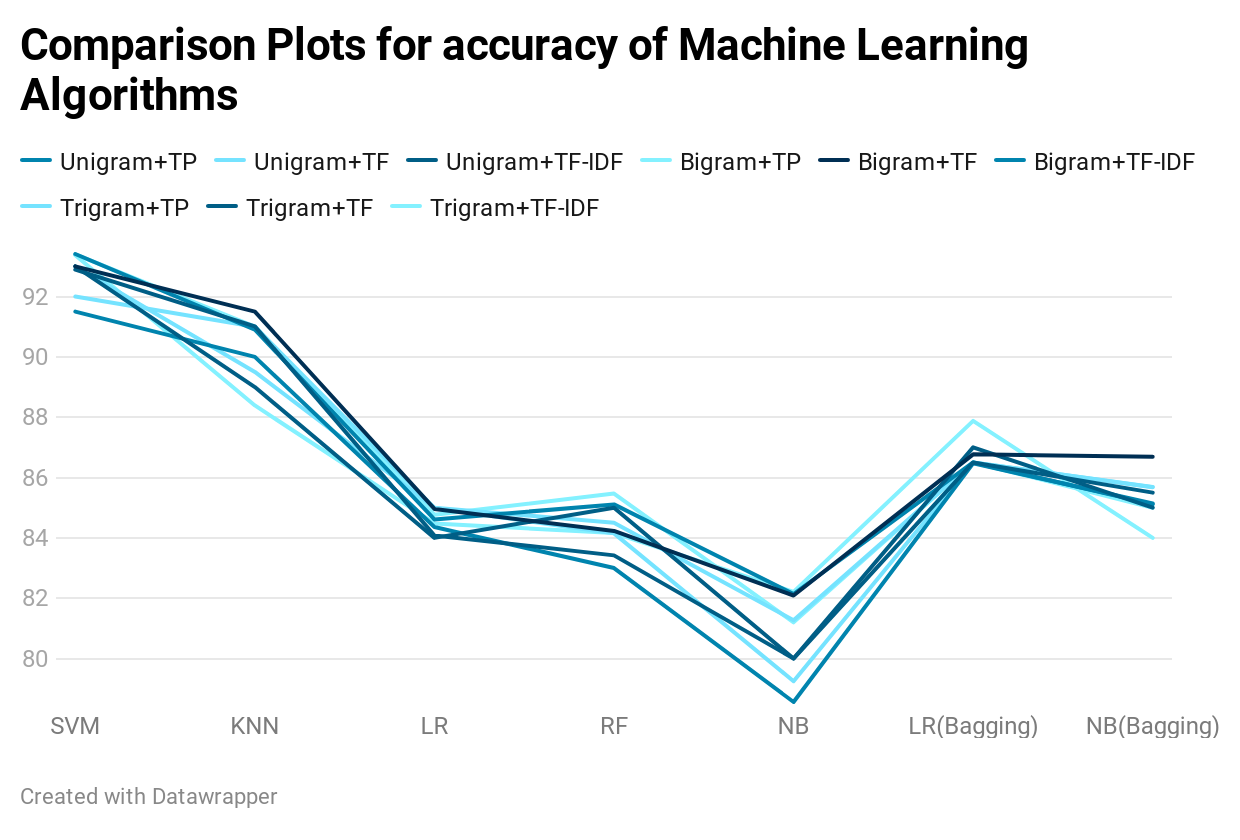}
         \subcaption{Machine Learning algorithms for Accuracy.}
           \label{f30}
 \end{minipage}
  \hfill
    \begin{minipage}[b]{0.5\textwidth}
      \includegraphics[width=\textwidth]{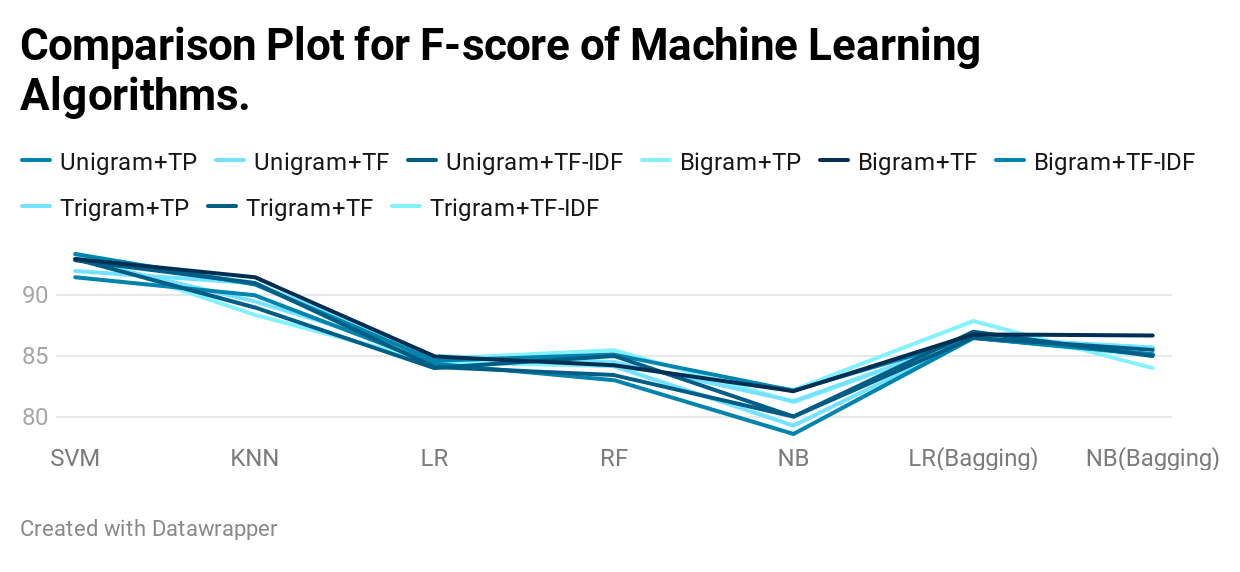}
        \subcaption{ Machine Learning algorithms for F-score.}
          \label{f31}
  \end{minipage}
  \hfill
  \begin{minipage}[b]{0.6\textwidth}
    \includegraphics[width=\textwidth]{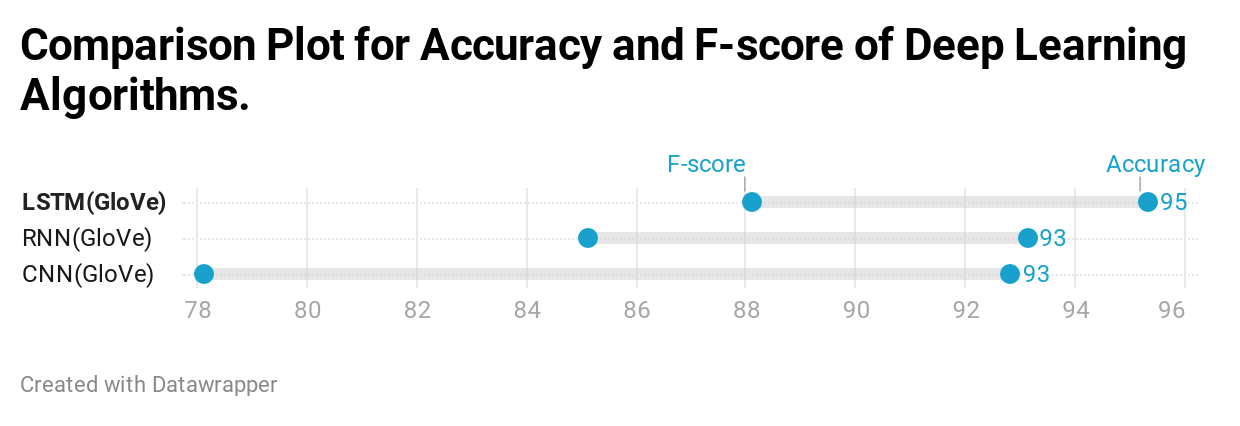}
     \subcaption{Deep Learning for Accuracy and F-score.}
      \label{f32}
  \end{minipage}
  
  \caption{ Visualization of Comparison Plot.}
\end{figure}

\begin{figure}
     \centering
     \begin{subfigure}[b]{0.4\textwidth}
         \centering
         \includegraphics[width=\textwidth]{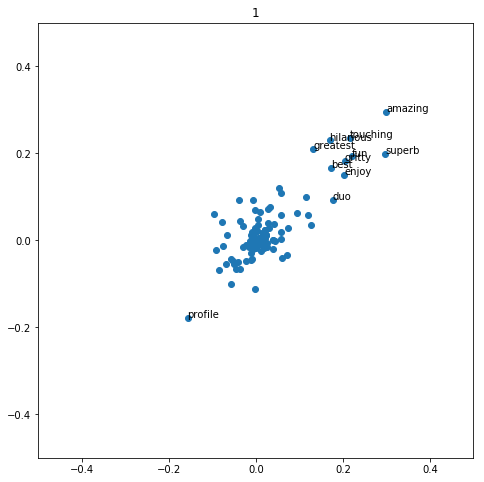}
         \caption{Positive words in reviews wrt Sentiment Polarity.}
         \label{f33}
     \end{subfigure}
     \hfill
     \begin{subfigure}[b]{0.4\textwidth}
         \centering
         \includegraphics[width=\textwidth]{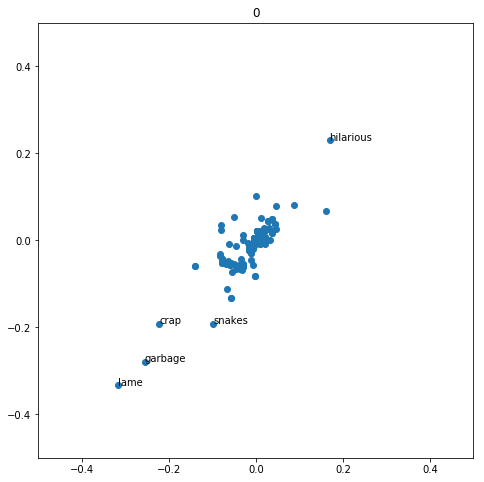}
         \caption{Negative words in reviews wrt Sentiment Polarity.}
         \label{f34}
     \end{subfigure}
     \hfill
     \caption{ Visualization of words through scatterplots. }
\end{figure}

\begin{figure}[!tbp]
  \centering
 \begin{minipage}[b]{0.3\textwidth}
    \includegraphics[width=\textwidth]{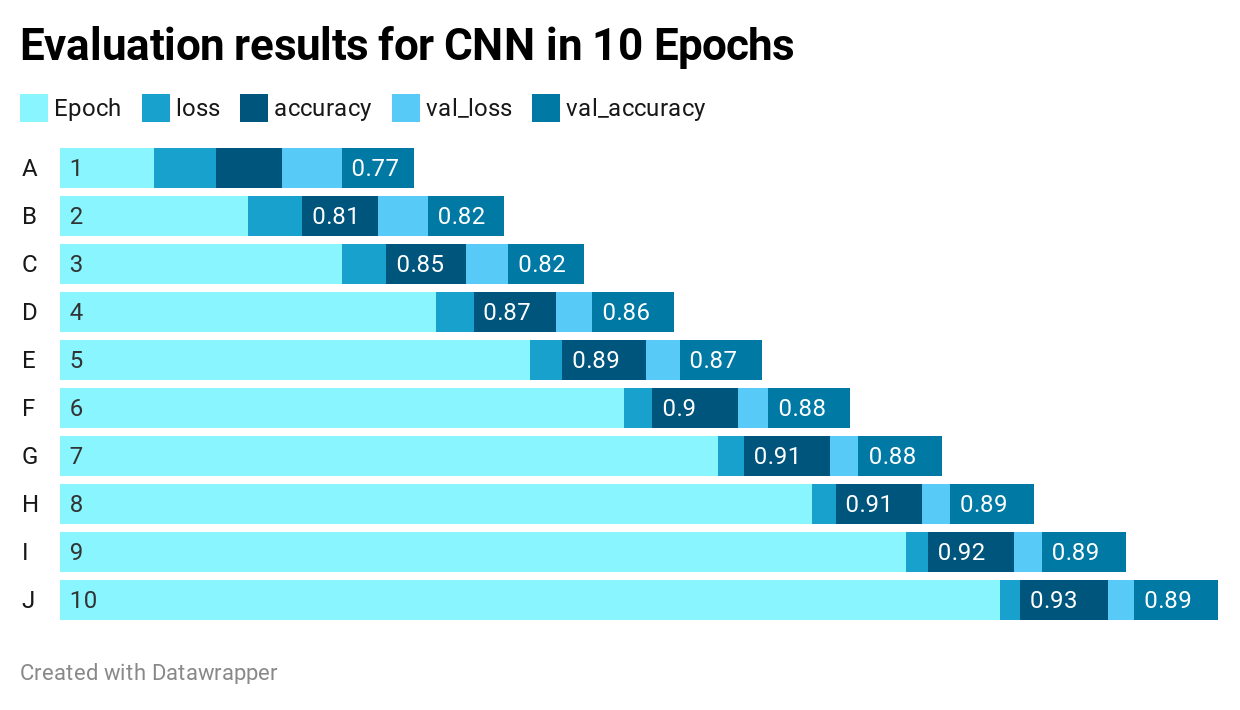}
     \subcaption{CNN plot for 10 Epochs.}
         \label{f35}
  \end{minipage}
  \hfill
  \begin{minipage}[b]{0.3\textwidth}
    \includegraphics[width=\textwidth]{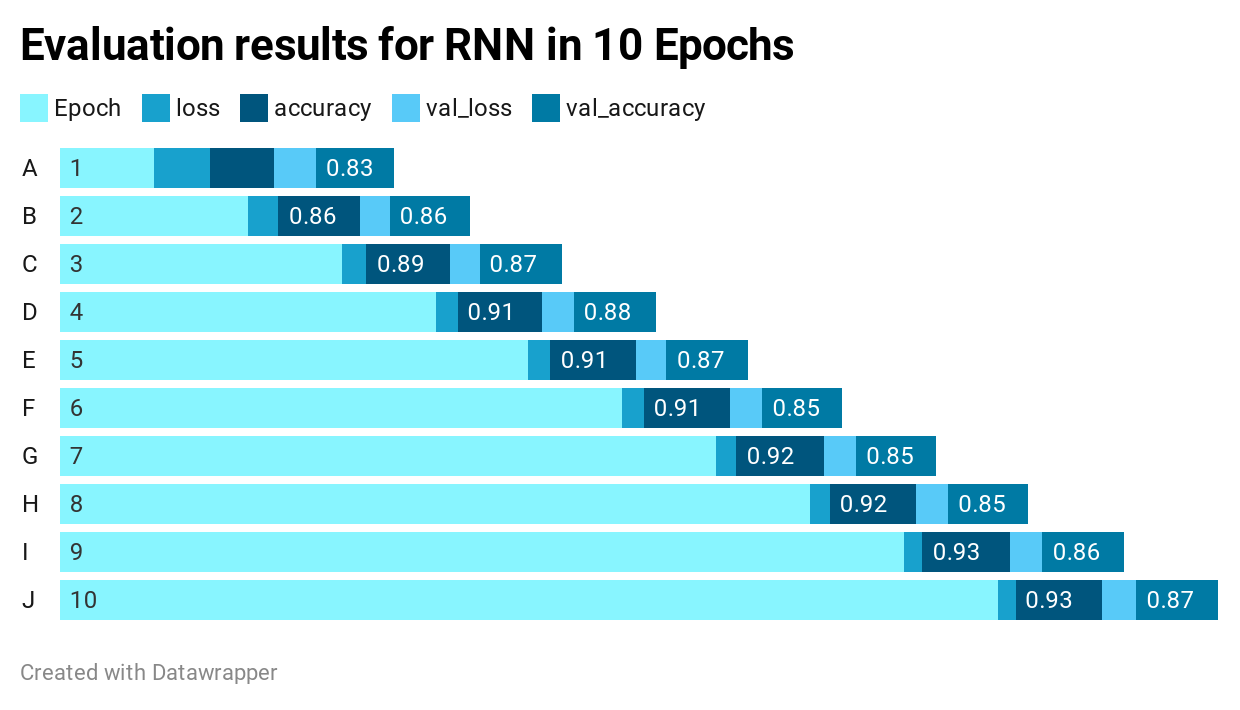}
       \subcaption{RNN plot for 10 Epochs.}
         \label{f36}
  \end{minipage}
\hfill
  \begin{minipage}[b]{0.3\textwidth}
    \includegraphics[width=\textwidth]{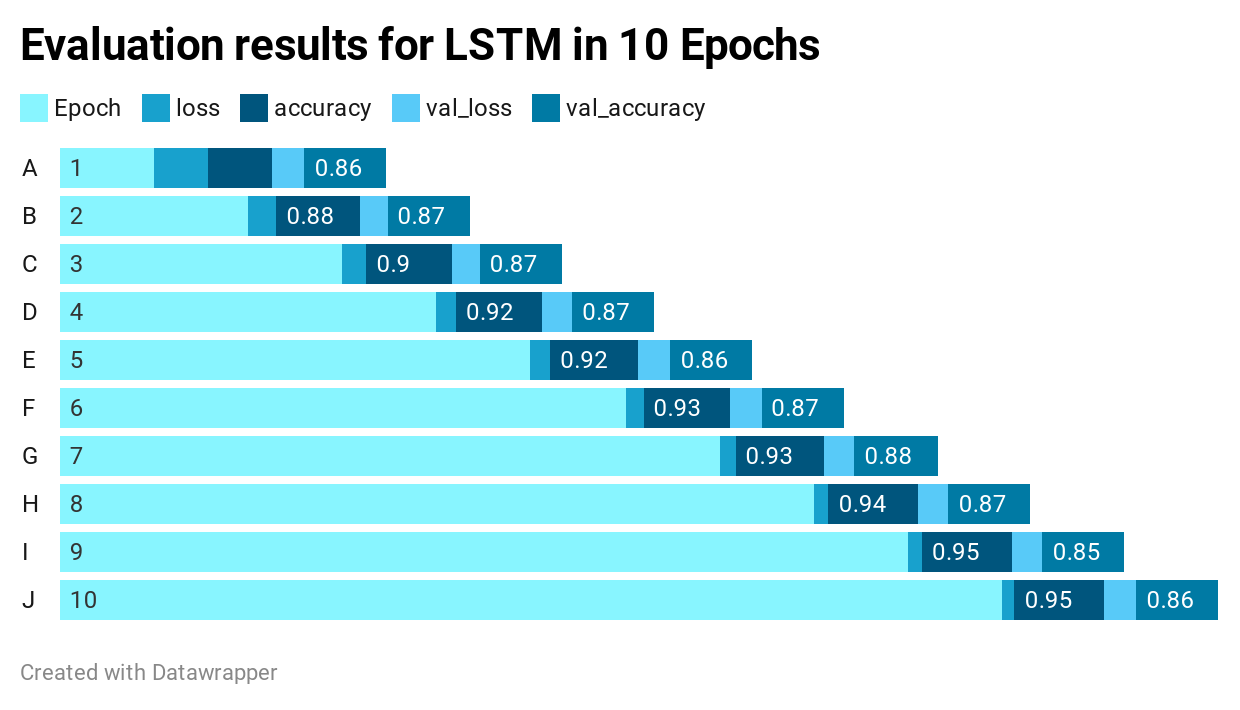}
     \subcaption{LSTM plot for 10 Epochs.}
         \label{f37}
  \end{minipage}
  \caption{ Visualization of Epochs in Deep Learning.}
\end{figure}

\begin{itemize}
    \item To encapsulate the key findings of deep learning algorithms on GloVe namely, LSTM, RNN, and CNN on the Google app reviews corpus and tested on SAR, we performed the experiments in 10 epochs in specified batch sizes. The accuracy and loss plots for the training dataset for LSTM, RNN, and CNN are depicted in Fig. 18a, 18b, 19a, 19b, 20a, 20b. respectively.
    
    \item The main findings in terms of loss and accuracy from the test dataset when applied with deep learning algorithms on GloVe are also plotted. These are listed in Fig. 18c, 18d, 19c, 19d, 20c, 20d. for LSTM, RNN, and CNN respectively.

    \item The accuracy for the test data is found out to be 97.8\% for LSTM, 89.5\% for RNN, and 85.9\% for CNN respectively.
    
    \item In our study, we have highlighted a pattern followed by the different conventional machine learning classifiers based on accuracy and F-score. Fig. 21a shows a comparison amongst the aforesaid machine learning algorithms based on accuracy via line chart.
    And Fig. 21b shows a comparison amongst the aforesaid machine learning algorithms based on F-scores via line chart.
    
    \item The main findings in terms of F-score and accuracy from the Google app reviews corpus when applied with deep learning algorithms in conjugation with GloVe are also plotted. The comparison plot is shown in Fig. 21c via a ranged scatter plot.
    
    \item From Table-6 of our study, we inferred, LSTM with GloVe attained the maximum accuracy. Using this, we modeled our corpus, Google app reviews, and showed a graphical analysis of Sentiment subjectivity vs Sentiment Polarity, over a specified category Entertainment, thereby picturing positive and negative words from the university students' reviews. These are illustrated through scatter plots in Fig. 22a, 22b.

    \item The empirical analysis of the deep learning paradigm, namely LSTM, RNN, CNN, in conjunction with GloVe was trained on the Google app reviews corpus, for 10 Epochs and tested on SAR. The key discoveries are sorted in descending order via a stacked plot in terms of Epochs and are shown in Fig. 23a, 23b, 23c respectively.
    
\end{itemize}

\subsection{Discussions}

\begin{itemize}
    \item From the graphical analysis, we answered 10 RQ's based on university students' viewpoints towards the app market when compared with the training data set. So, we could easily analyze the percentage of students with similarity in mindset and their inclination and liking and diversified knowledge towards the app market  
    
    \item From our study, the best algorithm LSTM along with word embedding, GloVe yielded the maximum accuracy of 95.2\% and an F-score of 0.88. This algorithm can also be tested for other word embedding namely FastText and Word2vec based on the same data set.
    
    \item Text analysis and machine learning techniques can foster the administrations of educational institutes to get feedback regarding the app market as used by university students. This might generate some learning about the valuable apps that could be made openly accessible to all students, if it is a paid one or if it is important to students for e-learning. 
    
    \item The presented text mining approach for sentiment analysis of university students' reviews crawled via a survey could also be initiated on a website, wherein university students of multiple universities within the same city could share their opinions on a common platform based on the commonly used or unique apps. Evaluating the e-learners reviews, identifying learners' emotions, based on text feedback in real-time sentiment analysis could be integrated with a deep learning-based framework.
    
    \item In the empirical analysis, conventional text representation schemes, ensemble methods, machine learning paradigms, and deep learning approaches have been considered.
    Ensemble learning techniques generated higher predictive performance when compared with conventional classification algorithms. From our study, a crystal clear inference is drawn that the deep learning model has outperformed the machine learning classifiers.
\end{itemize}

\section{Conclusions}

This work models the sentiment of the users using the Google reviews dataset and finds the university student's behavior towards the Google app market. Usually, the k-fold cross-validation technique is used for testing, i.e splitting the dataset in the ratio of 70:30 or 80:20. Not much research has been done in Sentiment analysis using students' reviews for testing. So, we had collected the real-life dataset from university students to study the proposed model. 
The exploratory analysis was initiated on our training dataset i.e., the Google app reviews. The results obtained were compared with the SAR through visualizations. 10 Research Questions were formulated and investigated to understand the correlation between the app market characteristics specifically price, popularity, sizing, categories, genres, and ratings of apps by the students when compared with that of the training data set. 
We have used a survey dataset in our study which is scalable, cost-effective, quick allows easy, and in depth analysis. A reader could easily explore the experiments by gathering data from across the globe. Applications of different Machine Learning and Deep Learning algorithms are seen with a close variation in results by exploring their pros and cons, and judging the misclassifications very minutely for an imbalanced dataset.


In this study, amongst the classification algorithm, SVM outperformed others in terms of accuracy(93.41\%) on the TF-IDF+bi-gram feature, while NB under-performed with an accuracy(78.56\%). In terms of F-score also, SVM outperformed other algorithms on TF-IDF on uni-gram, bi-gram schemes. KNN and LR performed significantly well and are fit for our data set. RF is also catching up. Bagging was implemented on LR and NB showing an apprehensive increment in accuracy and F-score. The corpus when trained using deep learning paradigms with word embedding namely GloVe showed that LSTM is highly suited for our study and some future research. It marked an outstanding accuracy of 95.2\%  and an F-score of 0.88. CNN and RNN performed averagely on GloVe with 93\% accuracy. In nutshell, the key takeaway of our study is that deep learning can be one of the grounds for further research on the university student dataset. A gap between opinions of the general public and students is pictorially and empirically illustrated well. Deep learning and machine learning algorithms are implemented efficiently.

\section{Limitations and Future Scope}
Despite having favorable results, there are certain limitations of this paper. 
To sum up, there were certain challenges faced while performing this study. They are:
\begin{itemize}
 
\item The collection of University students' reviews was indeed tedious and time-consuming. 
\item Moreover, the survey was localized only to a particular university and we did not get a 100\% participation of university students.
\item There were constraints on the reviews crawled from students, as they used abbreviations or short-form or SMS language, slang words, spelling mistakes, and emojis to express their reviews showing disbelief or disappointment.
\item Some students were reluctant to participate or gave false opinions or invaluable reviews and hence would create discrepancies as well.
    
\end{itemize}

For further improvement the future scope of our study could be listed as follows:
\begin{itemize}
    \item Empirical analysis of TP, TF, and TF-IDF based representation in conjugation with uni-gram, bi-gram, and tri-gram model respectively on other ensemble methods like random subspace and boosting. Hence, employing machine learning algorithms to study the predictive performance.
    
    \item BERT and XLNet could also be explored. 
    
   \item Analysis of word embedding namely, word2vec and FastText could also be explored on deep learning models (GRU and RNN-AM).
   
    \item Eventually, we could expand our data set by extending our online survey in other universities as well and more students within the range of the city. Some resource-crunched university students should be taken into account.
  
    \item Exploring the university students’ reviews in a multilingual domain and other resource-poor language could be encountered.
\end{itemize}

\section*{Acknowledgement}

We would like to dedicate this work of ours to one of our favourite Professors and former HoD Sir (late) Dr. B.K.Ratha, Department of Computer Science and Applications, Utkal University. A life full of endless possibilities cut short in a moment at an age of 54, due to deteriorating health. 

He encouraged us all the way and whose encouragement has made sure that "we give it all it takes to finish that which we have started". He is missed every day because a cruel twist of fate has left a void never to be filled in our lives. His care and concern for us knew no bounds. Our respect for him can never be quantified. May he find peace and happiness in Paradise!



\end{document}